\documentclass[letterpaper, 10 pt, conference]{ieeeconf}  

\IEEEoverridecommandlockouts                              
\makeatletter
\let\NAT@parse\undefined
\makeatother

\usepackage[dvipsnames]{xcolor}

\newcommand*\linkcolours{ForestGreen}

\usepackage{times}
\usepackage{graphicx}
\usepackage{amssymb}
\usepackage{gensymb}
\usepackage{amsmath}
\usepackage{breakurl}
\usepackage{caption}
\usepackage{subcaption}
\usepackage{cuted}

\usepackage[utf8]{inputenc}
\usepackage[ruled, linesnumbered]{algorithm2e} 
\usepackage[figuresright]{rotating}
\usepackage{booktabs} 
\usepackage{multirow} 

\usepackage{url,hyperref}
\hypersetup{
colorlinks,
linkcolor=\linkcolours,
citecolor=\linkcolours,
filecolor=\linkcolours,
urlcolor=\linkcolours}

\usepackage[labelfont={bf},font=small]{caption}
\usepackage[none]{hyphenat}

\usepackage{mathtools, cuted}

\usepackage[noadjust, nobreak]{cite}
\usepackage{tabularx}
\usepackage{amsmath}
\usepackage{float}
\usepackage{pifont}

\newcolumntype{Y}{>{\centering\arraybackslash}X}
\usepackage[]{placeins}

\usepackage{placeins}

\usepackage{tikz}

\usepackage[framemethod=tikz]{mdframed}
\usepackage{afterpage}
\usepackage{stfloats}
\usepackage{atbegshi}
\usepackage{amsmath}
\usepackage{graphicx}
\usepackage{subcaption}
\usepackage{multirow}
\newcommand{\handlethispage}{}
\newcommand{\discardpagesfromhere}{\let\handlethispage\AtBeginShipoutDiscard}
\newcommand{\keeppagesfromhere}{\let\handlethispage\relax}
\AtBeginShipout{\handlethispage}

\usepackage{comment}
\title{\LARGE \bf
Enhancing Neural Network Representations with Prior Knowledge-Based Normalization
}

\author{Bilal FAYE$^{1}$, Hanane AZZAG$^{2}$, Mustapha Lebbah$^{3}$, Djamel Bouchaffra$^{4}$  \\
	\normalsize e-mail: faye@lipn.univ-paris13.fr, azzag@univ-paris13.fr, mustapha.lebbah@uvsq.fr, dbouchaffra@cdta.dz
}

\begin{document}

\maketitle
\thispagestyle{empty}
\pagestyle{empty}

\begin{abstract}
Deep learning models face persistent challenges in training, particularly due to internal covariate shift and label shift. While single-mode normalization methods like Batch Normalization partially address these issues, they are constrained by batch size dependencies and limiting distributional assumptions. Multi-mode normalization techniques mitigate these limitations but struggle with computational demands when handling diverse Gaussian distributions. In this paper, we introduce a new approach to multi-mode normalization that leverages prior knowledge to improve neural network representations. Our method organizes data into predefined structures, or "contexts", prior to training and normalizes based on these contexts, with two variants: Context Normalization (CN) and Context Normalization - Extended (CN-X). When contexts are unavailable, we introduce Adaptive Context Normalization (ACN), which dynamically builds contexts in the latent space during training. Across tasks in image classification, domain adaptation, and image generation, our methods demonstrate superior convergence and performance. Our code implementation is available on our GitHub repository: \href{https://github.com/b-faye/prior-knowledge-norm}{https://github.com/b-faye/prior-knowledge-norm}.

\end{abstract}

\section{Introduction}
\label{introduction}
Deep Neural Networks (DNNs) are powerful models that apply stacked linear transformations with nonlinear activations~\cite{lecun2015deep}, enabling robust feature learning and representation but presenting significant training challenges, such as slow convergence, overfitting, and instability~\cite{glorot2010understanding}. The effectiveness of DNNs relies on advances in training methods that address these issues~\cite{hinton2006reducing, nair2010rectified, kingma2014adam, ioffe2015batch}.\newline
Normalization, a critical advancement, enhances training stability, optimizes learning, and improves generalization~\cite{ioffe2015batch, ulyanov2016instance, ioffe2017batch, wu2018group, ren2016normalizing, koccyigit2020unsupervised, luo2019switchable, luo2019mode, kalayeh2019training, liu2020evolving}. By reducing feature magnitude variations, normalization methods allow deeper layers to train efficiently, stabilizing activations for consistent layerwise distributions and accelerating convergence~\cite{lecun2002efficient}.\newline
Batch Normalization (BN), introduced by Ioffe and Szegedy~\cite{ioffe2015batch}, remains widely used, improving training efficiency by using batch-level statistics. However, BN’s reliance on batch size and its uniform data distribution assumption limit its applicability. To overcome these limitations, single-mode normalization methods address batch size dependencies~\cite{ba2016layer,ulyanov2016instance,ioffe2017batch,wu2018group,ren2016normalizing,koccyigit2020unsupervised}, while multi-mode approaches handle non-uniform data distributions~\cite{luo2019switchable,luo2019mode,kalayeh2019training,liu2020evolving}.\newline
In this paper, we introduce novel multi-mode normalization strategies that leverage prior knowledge to establish more accurate mini-batch data distributions. Our methods define distinct "contexts" within the input data—clusters of samples with shared characteristics—and use these predefined contexts to reduce the computational overhead typically associated with multi-mode normalization, which often requires dynamic mode estimation and additional parameters during training. We propose three variants: Context Normalization (CN), Context Normalization Extended (CN-X), and Adaptive Context Normalization (ACN). In CN and CN-X, normalization parameters are applied uniformly to samples within the same context in each mini-batch. Where context identification is challenging, ACN operates as a clustering-based approach, using only the desired number of contexts to dynamically normalize activations.\newline
Our methods achieve notable improvements in convergence and performance across image classification, domain adaptation, and image generation tasks.

\section{RELATED WORK}

\subsection{Single-mode normalization}
\label{state_normalization}
Single-mode normalization refers to normalization techniques that operate by standardizing activations using statistics computed from a single mode or source, such as a layer or mini-batch of data. These methods were pioneered by Batch Normalization (BN), introduced by Ioffe and Szegedy in their seminal work~\cite{ioffe2015batch}, which became a cornerstone of training neural networks.

\subsubsection{Batch Normalization Method}
BN normalizes activations by using the mean and variance calculated over mini-batches during training. This approach mitigates the problem of \textit{internal covariate shift}—the tendency of layer inputs to change distribution during training—thereby allowing higher learning rates and faster convergence. The normalization is done by centering the activations around zero with a mean of zero and scaling them with unit variance.\newline
Consider a 4-D activation tensor $\mathbf{x} \in \mathbb{R}^{N \times C \times H \times W}$ in a convolutional neural network, where $N$, $C$, $H$, and $W$ represent the batch size, number of channels, height, and width, respectively. BN computes the mini-batch mean ($\mu_B$) and standard deviation ($\sigma_B$) over the set $B = \{x_{1:m}: m \in [1, N] \times [1, H] \times [1, W]\}$, where $\mathbf{x}$ is flattened across all dimensions except the channel axis. A small constant $\epsilon$ is included for numerical stability, as shown in Equation~\ref{eq:1}.

\begin{equation}
    \mu_B = \frac{1}{m} \sum_{i=1}^m x_i \quad \sigma_B = \sqrt{\frac{1}{m} \sum_{i=1}^m (x_i - \mu_B)^2 + \epsilon}
    \label{eq:1}
\end{equation}

If the samples within the mini-batch come from the same distribution, the transformation $\mathbf{x} \rightarrow \mathbf{\hat{x}}$, as shown in Equation~\ref{eq:2}, produces a normalized distribution with zero mean and unit variance. BN then applies learnable scale ($\gamma$) and shift ($\beta$) parameters to re-scale the normalized data to a new distribution with mean $\beta$ and standard deviation $\gamma$.
\begin{equation}
        \hat{x}_i = \frac{x_i - \mu_B}{\sigma_B} \quad y_i = \gamma \hat{x}_i + \beta
        \label{eq:2}
\end{equation}
During inference, rather than using the batch statistics, BN employs a moving average of the mean and variance computed during training. The moving average mean $\bar{\mu}$ and variance $\bar{\sigma}^2$ are calculated as:

\begin{equation}
        \bar{\mu} = \alpha \bar{\mu} + (1 - \alpha) \mu_B \quad \bar{\sigma}^2 = \alpha \bar{\sigma}^2 + (1 - \alpha) \sigma_B^2
        \label{eq:bn_inference}
\end{equation}
Here, $\alpha$ is a momentum parameter that controls the update rate of the moving averages. During inference, these moving averages are used to normalize activations as:

\begin{equation}
        \hat{x}_i = \frac{x_i - \bar{\mu}}{\sqrt{\bar{\sigma}^2 + \epsilon}} \quad y_i = \gamma \hat{x}_i + \beta
        \label{eq:bn_transformation}
\end{equation}
This ensures consistency across different batch sizes during inference.\newline

Despite its remarkable performance in stabilizing the training of DNNs, BN faces significant limitations related to its dependency on mini-batch size. Specifically, BN's effectiveness diminishes when the size of the mini-batch is small. This occurs because BN relies on accurate estimates of batch statistics (mean and variance) during training, which become less reliable with smaller mini-batches, leading to noisy gradients and unstable updates. This limitation poses a challenge in scenarios where memory constraints or certain applications require smaller mini-batches. To address this issue, several variants of BN have been proposed.

\subsubsection{Extensions of Batch Normalization to Address Mini-Batch Dependency}
\label{chapter2:bn_extension}
To address the mini-batch dependency issue, several extensions of Batch Normalization have been introduced, including Layer Normalization (LN)~\cite{ba2016layer}, Instance Normalization (IN)~\cite{ulyanov2016instance}, Group Normalization (GN)~\cite{wu2018group}, and Divisive Normalization (DN)~\cite{ren2016normalizing}, Unsupervised Batch Normalization (UBN)~\cite{koccyigit2020unsupervised}. In this section, we adopt the notations from~\cite{kalayeh2019training,ren2016normalizing} to illustrate that the primary distinction between these methods lies in the specific set over which the sample statistics are computed.\newline
Let's consider $i=(i_N, i_C, i_L)$ as a vector indexing the tensor of activations $\mathbf{x} \in \mathbb{R}^{N\times C\times L}$, associated with a convolutional layer where the spatial domain has been flattened. The general normalization, $\mathbf{x} \rightarrow \mathbf{\hat{x}}$, is defined as:
\begin{equation}
    v_{i} = x_{i} - \mathbb{E}_{B_i}(\mathbf{x}), \quad
    \hat{x}_{i} = \frac{v_{i}}{\sqrt{\mathbb{E}_{B_i}(\mathbf{v}^2) + \epsilon}},
    \label{eq:general_transformation}
\end{equation}
where $\mathbb{E}_{B_i}(x)$ denotes the expectation computed over a subset $B_i$ of activations. Similar to BN, the normalized activations can be further adjusted by scaling and shifting using the parameters $\gamma$ and $\beta$. To derive the BN transformation (Equation~\ref{eq:bn_transformation}) from the general normalization Equation~\ref{eq:general_transformation}, it is only necessary to define the appropriate $B_i$ as:
\begin{equation}
        B_i = \{j: j_N \in [1, N], j_C \in [i_C], j_L \in [1, L]\}.
        \label{eq:bi_bn}
\end{equation}
In this case, $B_i$ captures all activations within the same channel $i_C$ across the entire mini-batch and spatial dimensions.\newline

\textbf{Layer Normalization (LN)}~\cite{ba2016layer} adapts BN for architectures like recurrent neural networks (RNNs), where temporal information is critical. Unlike BN, which normalizes across the mini-batch, LN normalizes across features for each training example independently, addressing RNN-specific challenges like varying batch sizes and dependencies on prior time steps. This ensures consistent normalization across all time steps, improving training stability and convergence. LN can be formulated as Equation~\ref{eq:general_transformation} when
\begin{equation}
        B_i = \{j: j_N \in [i_N], j_C \in [1, C], j_L \in [1, L]\}.
        \label{eq:bi_ln}
\end{equation}
LN effectively reduces internal covariate shift in RNNs, enhancing long-range dependency capture and performance in tasks like natural language processing and time-series forecasting. It's also computationally efficient and widely used in modern architectures like transformers~\cite{vaswani2017attention}. However, LN underperforms in convolutional layers, where local spatial variations are important, as it applies the same normalization across the entire spatial domain, making it less suited for convolutional architectures.\newline
\textbf{Instance Normalization (IN)}~\cite{ulyanov2016instance} extends the ideas of BN and LN, specifically designed for generative models and style transfer. Unlike BN, which normalizes across mini-batches, or LN, which normalizes across all features of a single example, IN normalizes each channel independently for each instance. This helps preserve instance-specific characteristics, making it particularly effective in tasks like image generation and style transfer, where separating content from style is crucial for creative manipulations and high-quality output~\cite{muVieCAST2023,inST2023}. IN can be formulated as Equation~\ref{eq:general_transformation} when
\begin{equation}
        B_i = \{j: j_N \in [i_N], j_C \in [i_C], j_L \in [1, L]\}.
        \label{eq:bi_in}
\end{equation}
However, IN can underperform in tasks like classification or CNN-based image recognition, where capturing correlations between instances is important. Its focus on instance-specific normalization can lead to a loss of shared statistics, limiting its effectiveness in scenarios that benefit from global feature interactions.\newline
\textbf{Group Normalization (GN)}~\cite{wu2018group} divides channels into smaller groups and computes the mean and variance for each group independently, making it robust to fluctuations in batch size. This is particularly useful in tasks like object detection and segmentation~\cite{bochkovskiy2020yolov4,carion2020end,chen2017deeplab,zhao2017pyramid}, where small batch sizes are common. GN balances the strengths of LN (G=1) and IN (G=C), providing more stable and effective normalization by ensuring group-specific statistics are representative of the data, leading to improved convergence and generalization. GN can be formulated as Equation~\ref{eq:general_transformation} when
\begin{equation}
        B_i = \{j: j_N \in [i_N], j_C \in [i_C], j_L \in [1, L] |  \lfloor \frac{j_C}{C/G} \rfloor\}
\end{equation}
However, GN's performance heavily depends on the choice of group size, requiring tuning to optimize results. While it outperforms BN in small-batch scenarios, it may underperform in very deep networks where capturing global batch statistics across all channels is crucial for effective feature learning.\newline
\textbf{Divisive Normalization (DN)}~\cite{ren2016normalizing} is a biologically inspired technique where each neuron's activity is divided by a weighted combination of its neighbors' activities, offering more dynamic control of activations. Unlike other methods that use simple statistics, DN adjusts activations as follows:
\begin{equation}
    v_{i} = x_{i} - \mathbb{E}_{A_i}(\mathbf{x}), \quad
\hat{x}_{i} = \frac{v_{i}}{\sqrt{\mathbb{E}_{B_i}(\mathbf{v}^2) + \rho^2}},
\end{equation}
where:
$$
A_i = \{j \mid d(x_i, x_j) \leq R_A \}, \quad B_i = \{j \mid d(v_i, v_j) \leq R_B\},
$$
with $d$ representing the distance between hidden units, $\rho$ the normalizer bias, and $R$ the neighborhood radius. This method enhances decorrelation of neuronal responses, reducing redundancy and improving focus on salient features. DN has shown to improve model robustness and interpretability, particularly in visual tasks. However, DN is computationally intensive, requiring the calculation of weighted sums for neighboring neurons, which can slow down large networks. Additionally, DN may underperform in convolutional networks, where global methods like BN are better at capturing broad data distributions. Its effectiveness also depends on fine-tuning parameters like neighborhood size and weights, adding complexity to model design. Thus, while DN has powerful benefits, its computational cost and complexity limit its broader use.\newline
\textbf{Unsupervised Batch Normalization (UBN)}~\cite{koccyigit2020unsupervised} addresses biased batch statistics in Batch Normalization (BN) when working with small labeled datasets. By incorporating additional unlabeled data from the same distribution to compute batch statistics, UBN reduces the bias introduced by small mini-batches. It is formulated as:
\begin{equation}
    B_i = \{j: j_N \in [1, i_N], j_C \in [i_C], j_L \in [1, L]\} \cup U_i,
\end{equation}
where $U_i$ represents the indices of unlabeled data. This approach enhances the representation of the data distribution, leading to more accurate normalization and stable training without needing changes to the network architecture. However, UBN relies on the assumption that the unlabeled data is from the same distribution as the labeled data; if there is a domain mismatch, the normalization may not generalize effectively.\newline 
\indent These techniques represent a significant step forward in overcoming the challenges of mini-batch dependency. Each method offers specific benefits suited to different DNN architectures and tasks. The choice of technique should be based on the model architecture and the training requirements, with newer methods providing a balance between flexibility and efficiency in training.

\subsection{Multi-mode normalization}
\label{chapter2:multi}
Multi-mode normalization standardizes activations using statistics from various sources, such as different layers, mini-batches, or feature channels. Several methods have been proposed to enhance this process, including Switchable Normalization (SwitchNorm)~\cite{luo2019switchable}, Mode Normalization (ModeNorm)~\cite{luo2019mode} and Mixture Normalization (MixNorm)~\cite{kalayeh2019training}. These techniques address the limitations of BN by overcoming the uniform data distribution assumption, which can hinder performance on diverse datasets. Overall, multi-mode normalization improves the robustness and stability of normalization in DNNs.\newline
\textbf{Switchable Normalization (SwitchNorm)}~\cite{luo2019switchable} is an advanced extension of BN that dynamically combines multiple normalization techniques, including BN, LN, and IN, through a set of learnable weights. Unlike BN, which assumes uniform data distribution across mini-batches and can suffer when batch sizes are small or when data distributions are not consistent, SwitchNorm allows the model to adaptively select the most appropriate normalization method for each layer. By leveraging this flexibility, SwitchNorm improves performance across a variety of scenarios, particularly when BN's reliance on mini-batch statistics becomes unreliable, such as in tasks with small batch sizes or non-uniform activations.\newline
For each activation $x_i$, SwitchNorm alters the normalization process by dynamically adjusting the computation of the batch statistics, as shown in Equation~\ref{eq:1}:
\begin{equation}
    \hat{x}_i = \frac{x_i - \sum_{k \in \Omega} w_k \mu_k}{\sqrt{\sum_{k \in \Omega} w'_k \sigma^2_k + \epsilon}}  \quad y_i = \gamma \hat{x}_i + \beta .
\end{equation}
Here, $\Omega$ represents a set of statistics estimated using different normalization methods. In the context of SwitchNorm, $\Omega = \{ \text{BN}, \text{LN}, \text{IN} \}$, which means that $\mu_k$ and $\sigma^2_k$ are computed for BN, LN, and IN using the batch $B_i$ as defined in Equations~\ref{eq:bi_bn}, \ref{eq:bi_ln}, and \ref{eq:bi_in} respectively. The calculations for these statistics can be expressed as follows:
\begin{equation}
    \mu_k = \frac{1}{|B_i|} \sum_{j \in B_i} x_j, \quad \sigma^2_k = \frac{1}{|B_i|} \sum_{j \in B_i} (x_j - \mu_k)^2.
\end{equation}
Furthermore, $w_k$ and $w'_k$ are importance ratios used to weight the means and variances, respectively. Each $w_k$ and $w'_k$ is a scalar variable constrained to the range $[0, 1]$, satisfying the conditions $\sum_{k \in \Omega} w_k = 1$ and $\sum_{k \in \Omega} w'_k = 1$. The weights $w_k$ can be computed as follows:
\begin{equation}
    w_k = \frac{e^{\lambda_k}}{\sum_{z \in \{ \text{BN}, \text{LN}, \text{IN} \}} e^{\lambda_z}}, \quad k \in \{ \text{BN}, \text{LN}, \text{IN} \},
\end{equation}
where $\lambda_{\text{BN}}, \lambda_{\text{LN}}$, and $\lambda_{\text{IN}}$ are control parameters learned during backpropagation. The weights $w'_k$ are defined similarly, using an additional set of control parameters $\lambda'_{\text{BN}}, \lambda'_{\text{LN}}, \lambda'_{\text{IN}}$.

Let $\Theta$ represent the set of network parameters (e.g., filters) and $\Phi$ denote the set of control parameters that define the network architecture. In SwitchNorm, the learned parameters are given by $\Phi = \{ \lambda_{\text{BN}}, \lambda_{\text{LN}}, \lambda_{\text{IN}}, \lambda'_{\text{BN}}, \lambda'_{\text{LN}}, \lambda'_{\text{IN}} \}$. Training a DNN with SwitchNorm involves minimizing the loss function:
$$
\min_{\{\Theta, \Phi\}} \frac{1}{N} \sum_{j=1}^{N} L(y_j, f(x_j; \Theta, \Phi)),
$$
where $\{x_j, z_j\}_{j=1}^N$ represents a set of training samples and their corresponding labels. The function $f(x_j; \Theta)$ is the model learned by the DNN to predict $z_j$. The parameters $\Theta$ and $\Phi$ are optimized jointly through backpropagation.\newline
SwitchNorm provides a valuable integration of various normalization methods but is limited by its dependence on BN, LN, and IN for parameter estimation. This reliance means it inherits the same limitations as these techniques, particularly in handling non-uniform data distributions, which may undermine its effectiveness in addressing the challenges posed by diverse data conditions.\newline
\textbf{Mode Normalization (ModeNorm)}~\cite{luo2019mode} introduces the concept of "modes" within the data. A mode refers to a dominant pattern or cluster within the data distribution, representing different subpopulations or variations in the input. ModeNorm detects these modes and normalizes the activations based on the statistics of their respective modes, rather than using the entire batch's statistics. This provides a more fine-grained and adaptive normalization process compared to SwitchNorm.\newline
For each activation $x_i$, ModeNorm adapts the normalization formula as follows:

\begin{equation}
\hat{x}_i = \sum_{k=1}^K g_k(x_i) \frac{x_i - \mu_k}{\sqrt{\sigma_k^2 + \epsilon}}  
\quad
y_i = \gamma \hat{x}_i + \beta,
\end{equation}
where $g_k, k \in \{1, ..., K\}$ are gating functions represented by a mixture of experts. For each $x_i$, $g_k(x_i) \in [0, 1]$ and $\sum_{k=1}^K g_k(x_i) = 1$. The estimators for $\mu_k$ and $\sigma^2_k$ are computed under the weighting from the gating network using the indices $B_i$:
\begin{equation}
\mu_k = \frac{1}{N_k} \sum_{j \in B_i} g_k(x_j) \cdot x_j \quad \sigma^2_k = \frac{1}{N_k} \sum_{j \in B_i} g_k(x_j) \cdot (x_j - \mu_k)^2,
\end{equation}
where $N_k = \sum_{j \in B_i} g_k(x_j)$. ModeNorm uses an affine transformation followed by softmax activation to represent the gating networks. When the number of modes $K = 1$, or when the gates collapse to a constant $g_k(x_i) = \text{const}$, ModeNorm reduces to BN. Like BN, during training, ModeNorm normalizes activations using statistics computed from the current batch. During inference, it uses moving averages of mean and variance, as in Equation~\ref{eq:bn_inference}, similarly to BN.\newline
ModeNorm helps overcome BN's shortcomings, especially when the data contains multiple modes or clusters that differ significantly. It excels in scenarios with non-uniform data distributions, where BN's global batch statistics may be misleading. However, ModeNorm adds complexity by requiring the identification of modes and calculating separate statistics for each mode, which can increase computational cost and introduce additional hyperparameters. Moreover, its effectiveness depends heavily on the accurate identification of modes, which may be challenging in complex or highly variable datasets, potentially limiting its generalizability in certain tasks.\newline
\textbf{Mixture Normalization (MixNorm)}~\cite{kalayeh2019training} extends BN by leveraging a probabilistic approach based on Gaussian Mixture Models (GMM). Rather than assuming a single underlying distribution for activations in a mini-batch, MixNorm captures the multimodal nature of data by normalizing each sample based on multiple modes. Each sample is assigned to one of several Gaussian components, enabling a more fine-grained adaptation of normalization to the underlying data distribution. This method improves on the limitations of BN, which can struggle with non-uniform or complex distributions across mini-batches.\newline
In MixNorm, the probability density function $p_\theta$ that characterizes the data is modeled as a GMM with $K$ components. The distribution for each sample $\mathbf{x} \in \mathbb{R}^D$ is expressed as:
\begin{equation}
\label{gmm}
    p(\mathbf{x}) = \sum_{k=1}^K \lambda_k p(\mathbf{x}|k), \quad \text{s.t.} \ \forall k: \lambda_k \geq 0, \ \sum_{k=1}^{K} \lambda_k = 1,
\end{equation}
where $\lambda_k$ is the mixture coefficient for the $k$-th component, and $p(\mathbf{x}|k)$ is the Gaussian distribution for the $k$-th component, given by:
\begin{equation}
    p(\mathbf{x}|k) = \frac{1}{(2\pi)^{D/2} |\Sigma_k|^{1/2}} \exp\left(-\frac{(\mathbf{x} - m_k)^T \Sigma_k^{-1} (\mathbf{x} - m_k)}{2}\right),
\end{equation}
with $m_k$ being the mean and $\Sigma_k$ the covariance matrix of the $k$-th Gaussian. Considering  $K$ components, MixNorm is implemented in two stages:
\begin{itemize}
    \item Estimation of the mixture model's parameters $\theta = \{\lambda_k, m_k, \Sigma_k: k = 1, \ldots, K\}$ using the Expectation-Maximization (EM) algorithm~\cite{Dempster77maximumlikelihood}.
    \item Normalization of each sample based on the estimated parameters and aggregation using posterior probabilities.
\end{itemize}
For a given activation $x_i$, the MixNorm transformation is formulated as:
\begin{equation}
    \label{mn_aggregation}
    \hat{x}_i = \sum_{k=1}^K \frac{p(k|x_i)}{\sqrt{\lambda_k}} \cdot \frac{x_i - \mu_k}{\sqrt{\sigma_k^2 + \epsilon}}, \quad y_i = \gamma \hat{x}_i + \beta,
\end{equation}
where $p(k|x_i) = \frac{\lambda_k p(x_i|k)}{\sum_{l=1}^K \lambda_l p(x_i|l)}$ represents the probability that $x_i$ belongs to the $k$-th component. The weighted mean and variance for the $k$-th component are computed as follows:
\begin{equation}
    \mu_k = \sum_j \frac{p(k|x_j)}{\sum_{l=1}^K p(l|x_j)} \cdot x_j,
\end{equation}
\begin{equation}
    \sigma^2_k = \sum_j \frac{p(k|x_j)}{\sum_{l=1}^K p(l|x_j)} \cdot (x_j - \mu_k)^2,
\end{equation}
MixNorm ensures that each sample is normalized according to the distribution it most likely belongs to, making it highly adaptive to complex, multimodal data distributions. MixNorm extends BN to heterogeneous complex datasets and often yield superior performance in supervised learning tasks. However, they are frequently computationally expensive due to the use EM algorithm.\newline

Activation normalization is a promising approach for addressing slow convergence and training instability in DNNs. BN, a single-mode method, has shown significant success by mitigating the internal covariate shift issue. However, BN's effectiveness diminishes when mini-batches are small or when the data samples within a batch come from different distributions. To address the small batch size problem, several single-mode alternatives such as LN, IN, GN, DN, and UBN have been introduced.\newline
To handle the challenge of non-uniform data distribution within mini-batches, multi-mode approaches such as SwitchNorm, ModeNorm, and MixNorm have been developed. However, this area is relatively underexplored, and existing methods tend to be computationally expensive, often requiring additional parameters or complex algorithms like EM in MixNorm. We propose three new multi-mode methods aimed at accelerating DNN training convergence and improving performance by leveraging prior knowledge-driven approaches.

\section{PROPOSED METHODS}
\label{proposed_methods}
Our contributions to DNN normalization techniques focus on accelerating convergence, enhancing stability, and boosting performance. We introduce multi-mode normalization strategies based on more accurate assumptions about mini-batch data distributions.\newline
Leveraging prior knowledge, our approach defines "contexts"—groups of samples with similar characteristics within the input data. Identifying these contexts before training reduces computational costs compared to traditional methods that dynamically estimate modes, often requiring extra parameters and resources.\newline
The structure of this section is as follows: Section~\ref{sec:prior_knowledge} introduces the concept of prior knowledge and various strategies for context construction. Section~\ref{sec:cn} presents Context Normalization (CN), Section~\ref{sec:cn_x} expands on Context Normalization Extended (CN-X), and Section~\ref{sec:acn} details Adaptive Context Normalization (ACN).
\subsection{Prior knowledge}
\label{sec:prior_knowledge}
In deep learning, prior knowledge—information or assumptions about the data or problem domain—guides the learning process and enhances efficiency by reducing the data needed for effective training~\cite{diligenti2017integrating, tobias1994interest, zhang2022deep, chen2020deep}. Starting from a more informed baseline, models focus on refinement rather than learning entirely from scratch, which improves generalization and mitigates overfitting, especially when labeled data is scarce~\cite{finn2017model, raissian2020learning}. Applications such as Knowledge-Driven Neural Networks (KD-NN) integrate human expertise, proving valuable in fields like healthcare~\cite{raissian2020learning}. Physics-Informed Neural Networks (PINN) embed physical laws within the model architecture, benefiting scientific simulations~\cite{finn2017model}. Similarly, knowledge graphs in NLP strengthen models' reasoning about entity relationships, improving tasks like semantic search~\cite{karpathy2019image}. By incorporating prior knowledge, deep learning models achieve greater data efficiency, robustness, and performance across diverse applications.\newline
In this study, we leverage prior knowledge by organizing input data into groups, or "contexts," which are defined using various strategies:
\begin{itemize}
    \item Clusters generated by algorithms such as k-means serve as contexts.
    \item Superclasses within datasets, grouping classes with similar characteristics, act as contexts.
    \item Domains in domain adaptation applications define contexts.
    \item In multimodal tasks, each modality operates as a distinct context.
\end{itemize}

\subsection{Context Normalization (CN)}
\label{sec:cn}
CN modifies Equation~\ref{mn_aggregation} from Mixture Normalization (MN)~\cite{kalayeh2019training}, where the mixture components are treated as modes for normalization. MN employs the Expectation-Maximization (EM) algorithm to estimate the parameters of these mixture components during training. However, EM is computationally expensive and must be applied repeatedly, as the activation distribution shifts with updates to the DNN weights.\newline
Instead of relying on the EM algorithm, we propose normalizing based on "contexts" that are pre-constructed from the input data before DNN training. Each sample in the input data is assigned to a single, unique context, with all samples within the same context sharing similar characteristics. Each sample belongs to a unique context $k$. CN ensures that all activations from a sample are associated with the same context $k$ throughout DNN training.\newline
To align with standard representations in the literature~\ref{state_normalization}, let $\mathbf{x} \in \mathbb{R}^{N \times C \times L}$ be an activation tensor, where $N$ is the batch size, $C$ is the number of channels, and $L = H \times W$ represents the flattened spatial dimensions (height $H$ and width $W$). Each activation is denoted by $\{x_i, k_i\}$, where $x_i$ is the activation and $k_i \in \{1, \dots, K\}$ is its context identifier, with $K$ being the number of contexts. Each activation $x_i$ is assigned to the context $k_i$ corresponding to the sample that produced it. Since each activation is associated with a unique known context, we have $p(k_i|x_i) = 1$ if $x_i$ belongs to context $k_i$, and $p(k_i|x_i) = 0$ otherwise. Consequently, Equation~\ref{mn_aggregation} simplifies to:
\begin{equation}
        \hat{x}_i = \frac{1}{\sqrt{\lambda_{k_i}}}.\frac{x_i - \mu_{k_i}}{\sqrt{\sigma_{k_i}^2 + \epsilon}} \quad y_i = \gamma_{k_i} \hat{x}_i + \beta_{k_i}
    \label{eq:scn_base}
\end{equation}
where $\lambda_{k_i}$ represents the proportion of samples in the dataset belonging to context $k_i$. The mean and variance are then defined as follows:
\begin{equation}
    \mu_{k_i} = \frac{1}{N_{k_i}} \cdot \sum_{x_i \in \mathbf{x}^{(k_i)}}  x_i
\end{equation}
\begin{equation}
    \sigma^2_{k_i} = \frac{1}{N_{k_i}} \cdot \sum_{x_i \in \mathbf{x}^{(k_i)}} (x_i - \mu_{k_i})^2
\end{equation}
where $\mathbf{x}^{(k_i)}$ denotes the subset of $\mathbf{x}$ containing the activations corresponding to context $k_i$, and $N_{k_i}$ represents the number of elements in $\mathbf{x}^{(k_i)}$. The moving averages of the mean $\bar{\mu}_{k_i}$ and variance $\bar{\sigma}^2_{k_i}$ are updated with a momentum rate $\alpha$ during training, following the same approach as in BN (see Equation~\ref{eq:bn_inference}). These updated statistics are then used to normalize the feature maps during inference:
\begin{equation}
    \bar{\mu}_{k_i} = \alpha\bar{\mu}_{k_i} + (1-\alpha)\mu_{k_i} \quad     \bar{\sigma}^2_{k_i} = \alpha\bar{\sigma}^2_{k_i} + (1-\alpha)\sigma^2_{k_i}
\end{equation}
In the special case where there is only a single context ($K = 1$), CN reduces to standard BN.\newline
\begin{algorithm}[!h]
\caption{CN Transform applied to activations of a specific context.}\label{alg:one}
\SetKwInOut{KwIn}{Input}
\SetKwInOut{KwOut}{Output}

\KwIn{$k$: context identifier;\\
      $\mathbf{x}^{(k)}$: subset of activations associated with context $k$;\\
      $\lambda_k$: proportion of samples in the dataset assigned to context $k$;\\
      $\{\gamma_k, \beta_k\}$: learnable parameters;
}
\KwOut{$\{y_i \} = \text{CN}_{\{\gamma_k, \beta_k\}}(k, \mathbf{x}^{(k)},\lambda_k)$
}
        $N_k = |\mathbf{x}^{(k)}|$ {\small \it // number of elements}\\
        $\mu_{k} = \frac{1}{N_{k}} \cdot \sum_{x_i \in \mathbf{x}^{(k)}}  x_i$ {\small \it // context mean}\\
        $\sigma^2_{k} = \frac{1}{N_{k}} \cdot \sum_{x_i \in \mathbf{x}^{(k)}} (x_i - \mu_{k})^2$ {\small \it // context variance}\\
        \For{$x_i \in \mathbf{x}^{(k)}$}{

            $\hat{x}_i = \frac{1}{\sqrt{\lambda_k}}.\frac{x_i - \mu_k}{\sqrt{\sigma_k^2 + \epsilon}}$ {\small \it // normalize}\\
            $y_i = \gamma_k \hat{x}_i + \beta_k $ {\small \it // scale and shift}\\
        }
\end{algorithm}
We present the CN transform (Algorithm~\ref{alg:one}), applied to a set of activations $\mathbf{x}^{(k)}$ of a specific context $k$. CN can be integrated into a network to manipulate activations. The scaled and shifted values $y = \{y_i\}$ are passed to other layers, while the normalized activations $\hat{x} = \{\hat{x}_i\}$, internal to CN, have mean $0$ and variance $1$. Unlike BN, which normalizes across the entire mini-batch, CN normalizes activations within context $k$. Each $\hat{x}$ is input to a sub-network with $y = \gamma_k \hat{x} + \beta_k$, accelerating training similarly to BN but per context $k$.\newline
During training, we need to propagate the gradient of loss $\ell$ through this transformation, as well as compute the gradients with respect to the parameters of CN transform. We use chain rule, as follows (before simplification):\newline
\begin{align*}
\frac{\partial \ell}{\partial \hat{x}_i} &= \frac{\partial \ell}{\partial y_i} \cdot \gamma_k \\
\frac{\partial \ell}{\partial \sigma_k^2} &= \frac{1}{\sqrt{\lambda_k}} \cdot \sum_{i=1}^{N_k} \frac{\partial \ell}{\partial \hat{x}_i} \cdot (x_i - \mu_k) \cdot \left(-\frac{1}{2}\right) \left(\sigma_k^2 + \epsilon\right)^{-\frac{3}{2}} \\
\frac{\partial \ell}{\partial \mu_k} &= \frac{1}{\sqrt{\lambda_k}} \cdot \left( \sum_{i=1}^{N_k} \frac{\partial \ell}{\partial \hat{x}_i} \cdot \frac{-1}{\sqrt{\sigma_k^2 + \epsilon}} \right) + \frac{\partial \ell}{\partial \sigma_k^2} \cdot \frac{\sum_{i=1}^{N_k} -2(x_i - \mu_k)}{N_k} \\
\frac{\partial \ell}{\partial x_i} &= \frac{\partial \ell}{\partial \hat{x}_i} \cdot \frac{1}{\sqrt{\lambda_k}} \cdot \frac{1}{\sqrt{\sigma_k^2 + \epsilon}} + \frac{\partial \ell}{\partial \sigma_k^2} \cdot \frac{2(x_i - \mu_k)}{N_k} + \frac{\partial \ell}{\partial \mu_k} \cdot \frac{1}{N_k} \\
\frac{\partial \ell}{\partial \gamma_k} &= \sum_{i=1}^{N_k} \frac{\partial \ell}{\partial y_i} \cdot \hat{x}_i \\
\frac{\partial \ell}{\partial \beta_k} &= \sum_{i=1}^{N_k} \frac{\partial \ell}{\partial y_i}
\end{align*}
The CN transform is a differentiable operation that introduces context-normalized activations into the network. This reduces internal covariate shift, accelerating training. Additionally, the learned affine transform, like in BN, allows CN to represent the identity transformation, preserving the network's capacity.\newline
\begin{algorithm}[!h]
\caption{Training a Context-Normalized Network.}\label{alg:two}
\SetKwInOut{KwIn}{Input}
\SetKwInOut{KwOut}{Output}

\KwIn{Net: Deep neural network with trainable parameters $\Theta$;\\
      $K$: number of contexts; \\
      $\{x_i, k_i\}$, where $k_i \in \{1, \dots, K\}$: activations and corresponding context;\\ 
      $\{\lambda_k\}_{k=1}^K$: proportion of samples assigned to each context $k$;\\
      $\{\gamma_k, \beta_k\}_{k=1}^K$: learnable parameters;\\
      $\alpha$: momentum;
}
\KwOut{Context-normalized network for inference, $\text{Net}^{\text{inf}}_{\text{CN}}$}
    $\text{Net}^{\text{tr}}_{\text{CN}} \gets \text{Net}$ {\small \it // Trainig CN deep neural network} \\
        \For{$k \gets 1$ to $K$}{
            Construct $\mathbf{x}^{(k)}$ with all activations for context $k$ \\
            Add transformation $y = \text{CN}_{\{\gamma_k, \beta_k\}}(k,\textbf{x}^{(k)},\lambda_k)$ to $\text{Net}^{\text{tr}}_{\text{CN}}$ (Algorithm~\ref{alg:one}) \\
            Replace the input $\mathbf{x}^{(k)}$ with $\mathbf{y}^{(k)}$ in each layer of $\text{Net}^{\text{tr}}_{\text{CN}}$\\
            $\bar{\mu}_k = \alpha\bar{\mu}_k + (1-\alpha)\mu_k \quad     \bar{\sigma}^2_k = \alpha\bar{\sigma}^2_k + (1-\alpha)\sigma^2_k$
        }
        Train $\text{Net}^{\text{tr}}_{\text{CN}}$ to optimize the parameters $\Theta \cup \{\gamma_k, \beta_k\}_{k=1}^K$\\
        $\text{Net}^{\text{inf}}_{\text{CN}} \gets \text{Net}^{\text{tr}}_{\text{CN}}$ {\small \it // Inference CN deep neural network with frozen parameters}\\
        \For{$k \gets 1$ to $K$}{
            Construct $\mathbf{x}^{(k)}$ with all activations for context $k$ \\
            \For{$x_i \in \mathbf{x}^{(k)}$}{
                $\hat{x}_i = \frac{1}{\sqrt{\lambda_k}}.\frac{x_i - \bar{\mu}_k}{\sqrt{\bar{\sigma}_k^2 + \epsilon}}$ {\small \it // normalize}\\
                $y_i = \gamma_k \hat{x}_i + \beta_k $ {\small \it // scale and shift}\\
            }
            Replace the input $\mathbf{x}^{(k)}$ with $\mathbf{y}^{(k)}$ in each layer of $\text{Net}^{\text{inf}}_{\text{CN}}$\\
        }
        
\end{algorithm}
To Context-Normalize a deep neural network, we define activations with their context identifiers $\{x_i, k_i\}$ and apply the CN transform on each based on its context, as outlined in Algorithm~\ref{alg:one}. Layers that previously received $\mathbf{x}^{(k)}$ (activations for context $k$) now take CN($k,\mathbf{x}^{(k)},\lambda_k$). This context-based normalization in mini-batches supports efficient training but isn't needed during inference; like BN, the output should depend deterministically on the input. After training, activations are normalized using:
\begin{equation}
\hat{y}_i = \gamma_{k_i} \cdot \frac{1}{\sqrt{\lambda_{k_i}}} \cdot \frac{x_i - \bar{\mu}_{k_i}}{\sqrt{\bar{\sigma}_{k_i}^2 + \epsilon}} + \beta_{k_i}
\end{equation}
Here, population statistics replace context-specific ones. Since the means and variances are fixed at inference, normalization reduces to a linear transform for each activation. This can be combined with the scaling by $\gamma_k$ and shift by $\beta_k$, resulting in a single linear transform replacing CN($k,\mathbf{x}^{(k)},\lambda_k$). Algorithm~\ref{alg:two} details the training process for context-normalized deep neural networks.

\textbf{Limitation.} CN divides the mini-batch into multiple subgroups based on predefined contexts, estimates the mean and variance for each subgroup, and normalizes the activations using the corresponding parameters. However, if a subgroup contains too few elements, the parameter estimates may become unreliable, causing CN to face the same issues as BN with small mini-batch sizes. To address this limitation, we propose an extension of CN, which we will discuss in Section~\ref{sec:cn_x}.

\subsection{Context Normalization Extended (CN-X)}
\label{sec:cn_x}
CN-X is an enhanced version of CN designed for more robust context parameter estimation. While CN estimates the normalization parameters (mean and variance) directly from activations within each context, CN-X instead learns these parameters as trainable weights of the neural network. These parameters are updated during backpropagation, making them more flexible and accurate over time. For each context $ k $, we define the parameter set $ \theta_k = \{\mu_k, \sigma_k^2\} $, where $ \mu_k $ and $ \sigma_k^2 $ are initialized randomly, with the constraint that $ \sigma_k^2 \geq 0 $.
\begin{algorithm}[!htbp]
\caption{CN-X Transform applied to activations of a specific context.}\label{alg:three}
\SetKwInOut{KwIn}{Input}
\SetKwInOut{KwOut}{Output}

\KwIn{$k$: context identifier;\\
      $\mathbf{x}^{(k)}$: subset of activations associated with context $k$;\\
      $\lambda_k$: proportion of samples in the dataset assigned to context $k$;\\
      $\phi_k = \{\mu_k, \sigma_k^2\}$ : normalization parameters;\\
      $\{\gamma_k, \beta_k\}$: learnable parameters;
}
\KwOut{$\{y_i \} = \text{CN-X}_{\{\phi_k, \gamma_k, \beta_k\}}(k, \mathbf{x}^{(k)},\lambda_k)$
}
        \For{$x_i \in \mathbf{x}^{(k)}$}{

            $\hat{x}_i = \frac{1}{\sqrt{\lambda_k}}.\frac{x_i - \mu_k}{\sqrt{\sigma_k^2 + \epsilon}}$ {\small \it // normalize}\\
            $y_i = \gamma_k \hat{x}_i + \beta_k $ {\small \it // scale and shift}\\
        }
\end{algorithm}
To normalize activations in context $k$, represented by $\mathbf{x}^{(k)}$, Algorithm~\ref{alg:one} is adapted to produce Algorithm~\ref{alg:three}. In this modified version, the normalization parameters $\theta_k$ are provided as inputs, and the normalization operation remains unchanged. However, unlike in CN, where the parameters are estimated directly from $\mathbf{x}^{(k)}$, in CN-X these parameters are learned through the network's training process.
\begin{algorithm}[!htbp]
\caption{Training a Context-Normalized Extended Network.}\label{alg:four}
\SetKwInOut{KwIn}{Input}
\SetKwInOut{KwOut}{Output}

\KwIn{Net: Deep neural network with trainable parameters $\Theta$;\\
      $K$: number of contexts; \\
      $\{x_i, k_i\}$, where $k_i \in \{1, \dots, K\}$: activations and corresponding context;\\ 
      $\{\lambda_k\}_{k=1}^K$: proportion of samples assigned to each context $k$;\\
      $\{\gamma_k, \beta_k\}_{k=1}^K$: learnable parameters;\\
      $\alpha$: momentum;
}
\KwOut{Context-normalized Extended network for inference, $\text{Net}^{\text{inf}}_{\text{CN-X}}$}
    Random initialize $\phi_k = \{\mu_k, \sigma^2_k\}$, where $k \in \{1, ..., K\}$ {\small \it // initialize normalization parameters}\\
    $\text{Net}^{\text{tr}}_{\text{CN-X}} \gets \text{Net}$ {\small \it // Trainig CN-X deep neural network} \\
        \For{$k \gets 1$ to $K$}{
            Construct $\mathbf{x}^{(k)}$ with all activations for context $k$ \\
            Add transformation $y = \text{CN-X}_{\{\phi_k, \gamma_k, \beta_k\}}(k,\textbf{x}^{(k)},\lambda_k)$ to $\text{Net}^{\text{tr}}_{\text{CN-X}}$ (Algorithm~\ref{alg:one}) \\
            Replace the input $\mathbf{x}^{(k)}$ with $\mathbf{y}^{(k)}$ in each layer of $\text{Net}^{\text{tr}}_{\text{CN-X}}$\\
        }
        Train $\text{Net}^{\text{tr}}_{\text{CN-X}}$ to optimize the parameters $\Theta \cup \{\phi_k, \gamma_k, \beta_k\}_{k=1}^K$\\
        $\text{Net}^{\text{inf}}_{\text{CN-X}} \gets \text{Net}^{\text{tr}}_{\text{CN-X}}$ {\small \it // Inference CN-X deep neural network with frozen parameters}\\
        \For{$k \gets 1$ to $K$}{
            Construct $\mathbf{x}^{(k)}$ with all activations for context $k$ \\
            \For{$x_i \in \mathbf{x}^{(k)}$}{
                $\hat{x}_i = \frac{1}{\sqrt{\lambda_k}}.\frac{x_i - \mu_k}{\sqrt{\sigma_k^2 + \epsilon}}$ {\small \it // normalize}\\
                $y_i = \gamma_k \hat{x}_i + \beta_k $ {\small \it // scale and shift}\\
            }
            Replace the input $\mathbf{x}^{(k)}$ with $\mathbf{y}^{(k)}$ in each layer of $\text{Net}^{\text{inf}}_{\text{CN-X}}$\\
        }
        
\end{algorithm}
Algorithm~\ref{alg:four} outlines the process for training a neural network with CN-X. Let $\Theta$ represent the neural network parameters, and $\Phi = \{\phi_k\}_{k=1}^K$, where $\phi_k = \{\mu_k, \sigma_k^2\}$, denote the set of learnable normalization parameters. The objective is to minimize the loss function:
$$
\min_{\Theta, \Phi} \frac{1}{N} \sum_{j=1}^N \ell(z_j, f(x_j; \Theta, \Phi)),
$$
where $\{x_j, z_j\}_{j=1}^N$ is the set of training samples and labels, with each sample belonging to a single context $k_j \in \{1, \dots, K\}$. The function $f(x_j; \Theta, \Phi)$ is learned by the network to predict the output $y_j$. Both $\Theta$ and $\Phi$ are optimized jointly via backpropagation.\newline
This approach differs from previous methods like BN and CN, where normalization parameters $\Phi$ are often treated as separate network modules (e.g., scale and shift) and not essential for normalization. In CN-X, $\Phi$ is learned directly during training, contributing to minimizing the loss function. Since the normalization parameters are not estimated from the activations, even small context sizes in a mini-batch do not negatively impact the learned parameters, as they are updated as part of the network's weights.\newline
Similar to CN, in CN-X, we need to propagate the gradient of the loss function $\ell$ through the transformation during training, while also computing the gradients with respect to the parameters of the CN-X transformation. This is achieved by applying the chain rule, as outlined below (prior to simplification):\newline
\begin{align*}
\frac{\partial \ell}{\partial \hat{x}_i} &= \frac{\partial \ell}{\partial y_i} \cdot \gamma_k \\
\frac{\partial \ell}{\partial \sigma_k^2} &= \frac{1}{\sqrt{\lambda_k}} \cdot \sum_{i=1}^{N_k} \frac{\partial \ell}{\partial \hat{x}_i} \cdot (x_i - \mu_k) \cdot \left(-\frac{1}{2}\right) \left(\sigma_k^2 + \epsilon\right)^{-\frac{3}{2}} \\
\frac{\partial \ell}{\partial \mu_k} &= \frac{1}{\sqrt{\lambda_k}} \cdot \left( \sum_{i=1}^{N_k} \frac{\partial \ell}{\partial \hat{x}_i} \cdot \frac{-1}{\sqrt{\sigma_k^2 + \epsilon}} \right) \\
\frac{\partial \ell}{\partial x_i} &= \frac{\partial \ell}{\partial \hat{x}_i} \cdot \frac{1}{\sqrt{\lambda_k}} \cdot \frac{1}{\sqrt{\sigma_k^2 + \epsilon}} \\
\frac{\partial \ell}{\partial \gamma_k} &= \sum_{i=1}^{N_k} \frac{\partial \ell}{\partial y_i} \cdot \hat{x}_i \\
\frac{\partial \ell}{\partial \beta_k} &= \sum_{i=1}^{N_k} \frac{\partial \ell}{\partial y_i}
\end{align*}
\textbf{Limitations.} CN-X methods rely on predefined contexts within the input dataset for normalization. In domains where constructing these contexts is challenging, such approaches become difficult to apply effectively. To address this limitation, we propose Adaptive Context Normalization (ACN), a method that retains the strengths of both CN-X and CN without the need for predefined contexts. We will elaborate on ACN in Section~\ref{sec:acn}.

\subsection{Adaptive Context Normalization (ACN)}
\label{sec:acn}
In ACN, we shift our focus from predefining contexts within the input dataset to dynamically constructing them during the training of the neural network. Unlike CN-X and CN, where inputs are represented as $(x_i, k_i)$—indicating predefined contexts—ACN simplifies this representation to just $x_i$. ACN only requires the specification of the number of contexts, $K$, to be created during the normalization process, akin to clustering algorithms that use a predefined number of clusters. However, instead of relying on prior knowledge or fixed clusters, ACN allows the neural network to autonomously discover a latent space of activations that adheres to a GMM. During training, ACN incrementally clusters neuron activations without predefined partitions, enabling the model to adapt to task-specific challenges without prior cluster information. This flexibility permits the neural network to explore and adapt to the underlying patterns in the data independently. Since the specific context for each activation is not predetermined, ACN utilizes Equation~\ref{mn_aggregation} to normalize across all contexts. Unlike traditional methods such as MN, where parameters are often fixed, ACN learns the parameters of these contexts as neural network weights during backpropagation. This approach eliminates the need for computationally intensive algorithms like EM, enhancing efficiency in the training process.\newline
The GMM parameters $\theta = \{\lambda_k, m_k, \Sigma_k: k = 1, \ldots, K\}$ are optimized in alignment with the target task. Algorithm~\ref{alg:five} outlines the training procedure of a deep neural network using ACN as the normalization method. Initially, the GMM parameters are randomly initialized, ensuring that $\sum_{k=1}^K \lambda_k = 1$ is maintained throughout training. This integration allows the GMM parameter estimation to become a dynamic part of the neural network, offering a more adaptive approach. Unlike methods like MN that rely on the EM algorithm—which cannot efficiently track changes in the activation distribution due to its high computational cost—this approach continuously updates the GMM parameters based on shifts in the activation distribution.
\begin{algorithm}[!htbp]
\caption{Training a Adaptive Context-Normalized Network.}\label{alg:five}
\SetKwInOut{KwIn}{Input}
\SetKwInOut{KwOut}{Output}

\KwIn{Net: Deep neural network with trainable parameters $\Theta$;\\
      $K$: number of contexts; \\
      $\{x_i\}$: set of activations;\\ 
      $\{\gamma_k, \beta_k\}_{k=1}^K$: learnable parameters;\\
      $\alpha$: momentum;
}
\KwOut{Context-normalized Extended network for inference, $\text{Net}^{\text{inf}}_{\text{ACN}}$}
        Initialize the parameters for each context as follows: \newline
        $\theta_k = \{\lambda_k$, $\mu_k$, $\Sigma_k\}$ for $k \in \{1, ..., K\}$, subject to the condition that
        $\sum_{k=1}^K \lambda_k = 1$\\
        
        \For{$x_i \in \mathbf{x}$}{
            Add transformation $\hat{x}_i$ using Equation~\ref{mn_aggregation} \\
            Modify each layer in $Net^{tr}_{ACN}$ with input $x_i$ to take $\hat{x}_i$ instead \\
            
        }
        
        Train $\text{Net}^{\text{tr}}_{\text{ACN}}$ to optimize the parameters $\Theta \cup \{\theta_k, \gamma_k, \beta_k\}_{k=1}^K$\\
        $\text{Net}^{\text{inf}}_{\text{ACN}} \gets \text{Net}^{\text{tr}}_{\text{ACN}}$ {\small \it // Inference ACN deep neural network with frozen parameters}\\
            \For{$x_i \in \mathbf{x}$}{
                $\hat{x}_{i} = \sum_{k=1}^K \frac{p(k|x_i)}{\sqrt{\lambda_k}} \left(\frac{x_i-\mu_k}{\sqrt{\sigma^2_k + \epsilon}}\right)$ {\small \it // normalize}\\
                $y_i = \gamma_k \hat{x}_i + \beta_k $ {\small \it // scale and shift}\\
            }
            Replace the input $\mathbf{x}$ with $\mathbf{y}$ in each layer of $\text{Net}^{\text{inf}}_{\text{ACN}}$\\        
\end{algorithm}
As the two approaches (CN and CN-X), in ACN we need to propagate the gradient of the loss function $\ell$ through the transformation during training. This is achieved by applying the chain rule, as outlined below (prior to simplification):
\small 
\begin{align*}
\frac{\partial \ell}{\partial \hat{x}_i} &= \frac{\partial \ell}{\partial y_i} \cdot \gamma_k \\
\frac{\partial \ell}{\partial \sigma_k^2} &= - \frac{1}{2(\sigma_k^2 + \epsilon)^{3/2}} \sum_{i=1}^N \frac{\partial \ell}{\partial x_i} \cdot \frac{p(k|x_i)}{\sqrt{\lambda_k}} \cdot (x_i - \mu_k) \\
\frac{\partial \ell}{\partial \mu_k} &= - \sum_{i=1}^N \frac{\partial \ell}{\partial x_i} \cdot \frac{p(k|x_i)}{\sqrt{\lambda_k}} \cdot \frac{1}{\sqrt{\sigma_k^2 + \epsilon}} \\
& \quad + \frac{\partial \ell}{\partial \sigma_k^2} \cdot \left( -2 \sum_{i=1}^N \frac{p(k|x_i)}{\sum_{l=1}^K p(l|x_i)} \cdot (x_i - \mu_k) \right) \\
\frac{\partial \ell}{\partial p(k|x_i)} &= \frac{\partial \ell}{\partial \hat{x}_i} \cdot \frac{1}{\lambda_k} \cdot \frac{x_i - \mu_k}{\sqrt{\sigma^2_k + \epsilon}} \\
& \quad + \frac{\partial \ell}{\partial \sigma^2_k} \cdot \frac{\sum_{l=1}^K p(l|x_i) - p(k|x_i)}{(\sum_{l=1}^K p(l|x_i))^2} \\
& \quad + \frac{\partial \ell}{\partial \mu_k} \cdot \frac{\sum_{l=1}^K p(l|x_i) - p(k|x_i)}{(\sum_{l=1}^K p(l|x_i))^2}\cdot x_i \\
\frac{\partial \ell}{\partial p(x_i|k)} &= \frac{\partial \ell}{\partial p(k|x_i)} \cdot \frac{\lambda_k (\sum_{l=1}^K p(x_i|l) - p(x_i|k))}{(\sum_{l=1}^K p(x_i|l))^2} \\
\frac{\partial \ell}{\partial m_k} &= \sum_{i=1}^N \frac{\partial \ell}{\partial p(x_i|k)} \cdot \frac{1}{(2\pi)^{D/2}|\Sigma_k|^{1/2}} \\
& \quad \cdot (\Sigma_k^{-1}(x_i - m_k))\exp\left(\frac{(x_i - m_k)^T\Sigma_k^{-1}(x_i - m_k)}{2}\right)\\
\frac{\partial \ell}{\partial \Sigma_k} &= \sum_{i=1}^N \frac{\partial \ell}{\partial p(x_i|k)} \cdot \frac{1}{(2\pi)^{D/2}|\Sigma_k|^{1/2}} \\
& \quad \cdot \left(\frac{1}{2} \Sigma_k^{-1}(x_i - m_k)(x_i - m_k)^T\Sigma_k^{-1}\right)\\
& \quad \cdot \exp\left(\frac{(x_i - m_k)^T\Sigma_k^{-1}(x_i - m_k)}{2}\right)\\
\frac{\partial \ell}{\partial x_i} &= \frac{\partial \ell}{\partial \hat{x}_i} \cdot \sum_{k=1}^K \frac{p(k|x_i)}{\sqrt{\lambda_k}} \cdot \frac{1}{\sqrt{\sigma^2_k + \epsilon}} \\
& \quad + \frac{\partial \ell}{\partial \sigma^2_k} \cdot \frac{p(k|x_i)}{\sum_{l=1}^K p(l|x_i)} \cdot 2(x_i - \mu_k) + \frac{\partial \ell}{\partial p(x_i|k)} \\
& \quad + \frac{\partial}{\partial \mu_k}  \cdot \frac{p(k|x_i)}{\sum_{l=1}^K p(l|x_i)} \frac{1}{(2\pi)^{D/2}|\Sigma_k|^{1/2}} \\
& \quad \cdot (-\Sigma_k^{-1}(x_i - m_k))\exp\left(-\frac{1}{2}(x_i - m_k)^T\Sigma_k^{-1}(x_i - m_k)\right)\\
& \quad + \frac{\partial \ell}{\partial p(x_i|k)} \\
& \quad \cdot \frac{\lambda_k\left[\frac{\partial p(x_i|k)}{\partial x_i}\sum_{l=1}^K \lambda_k p(x_i|l) - p(x_i|k)\sum_{l=1}^K \lambda_l \frac{\partial p(x_i|l)}{\partial x_i}\right]}{(\sum_{l=1}^K \lambda_l p(x_i|l))^2}\\
\frac{\partial \ell}{\partial \gamma_k} &= \sum_{i=1}^{N} \frac{\partial \ell}{\partial y_i} \cdot \hat{x}_i \\
\frac{\partial \ell}{\partial \beta_k} &= \sum_{i=1}^{N} \frac{\partial \ell}{\partial y_i}
\end{align*}
\normalsize
The ACN is a differentiable operation that integrates context-sensitive, normalized activations directly into the neural network. This method is particularly advantageous for scenarios involving multi-modal data distributions, as it unifies normalization across multiple modes without requiring complex, separate algorithms for estimating mode-specific parameters. Instead, ACN dynamically adapts its normalization based on context.\newline
In this approach, we use MN as a baseline; however, ACN is not limited to MN and can be generalized to other normalization techniques, such as ModeNorm. The key advantage lies in how ACN enables the model to learn context-relevant parameters, which adapt based on the activation distributions that shift throughout training as the network's weights are updated via backpropagation.\newline
By leveraging adaptive context normalization, the method allows for smoother transitions and better performance across different data modes, ensuring more efficient parameterization without the need for additional heavy computations during training. This flexibility makes ACN an appealing approach for tasks where data has varying distributions or requires context-sensitive handling.

\section{EXPERIMENTATION}
In this section, we present several applications where we compare traditional normalization techniques (see Section~\ref{state_normalization}) with our proposed normalization methods (see Section~\ref{proposed_methods}). These comparisons are demonstrated across various tasks, including image classification (Section~\ref{image_classification}), domain adaptation (Section~\ref{domain_adaptation}), and image generation (Section~\ref{image_generation}). We utilize several well-known benchmark datasets that are widely recognized within the classification community:
\begin{itemize}
    \item \textbf{CIFAR-10:} A dataset with 50,000 training images and 10,000 test images, each of size $32\times32$ pixels, distributed across 10 classes~\cite{cifar10_datasets}.
    \item \textbf{Oxford-IIIT Pet:} A dataset containing images of 37 different breeds of cats and dogs, with approximately 200 images per breed~\cite{oxford_pets_dataset}. 
    \item \textbf{CIFAR-100:} Derived from the Tiny Images dataset, it consists of 50,000 training images and 10,000 test images of size $32\times32$, divided into 100 classes grouped into 20 superclasses~\cite{cifar100_datasets}.
    \item \textbf{Tiny ImageNet:} A dataset that is a reduced version of the ImageNet dataset~\cite{deng2009imagenet}, containing 200 classes with 500 training images and 50 test images per class~\cite{le2015tiny}.
    \item \textbf{MNIST digits:} Contains 70,000 grayscale images of size $28\times28$ representing the 10 digits, with around 6,000 training images and 1,000 testing images per class~\cite{mnist_datasets}.
    \item \textbf{SVHN:} A challenging dataset with over 600,000 digit images, focusing on recognizing digits and numbers in natural scene images~\cite{sermanet2012convolutional}.
\end{itemize}
For applying CN and CN-X, we will use three approaches to build contexts: (i) applying the k-means algorithm to create clusters and using these clusters as contexts, (ii) utilizing superclasses, which are groups of classes, as contexts, and (iii) treating each domain in domain adaptation as a separate context.

To evaluate our normalization methods (CN, CN-X, and ACN) against traditional normalization techniques (BN, LN, MixNorm, and ModeNorm) in image classification tasks, we employ the DenseNet architecture~\cite{huang2017densely}, varying the number of layers to create two distinct models: a shallow model with 40 layers (DenseNet-40) and a deeper model with 100 layers (DenseNet-100).\newline
DenseNet employs BN as the normalization layer. We create a corresponding DenseNet model for each normalization technique (LN, MixNorm, ModeNorm, CN, CN-X, and ACN) by replacing the BN layers with the specific normalization method.\newline
In the first experiment, detailed in the section "Building Custom Contexts", we will employ the k-means algorithm to generate clusters that will act as contexts for CN and CN-X, utilizing the Oxford IIIT Pet, CIFAR-10, CIFAR-100, and Tiny ImageNet datasets. In the second experiment, outlined in the section "Leveraging Predefined Contexts", we will utilize the superclass structure (groups of classes) within the Oxford-IIIT Pet and CIFAR-100 datasets as contexts.
\subsubsection{Building Custom Contexts}
\label{costum_context}
In this study, we assume that the underlying structure of the dataset is not well understood, and there is no clear prior knowledge regarding the contextual relationships within the data. To address this, we need to establish these contexts before training our neural networks, specifically DenseNet-40 and DenseNet-100, for both CN and CN-X normalization techniques. To define the contexts, we employ the k-means clustering algorithm, treating the resulting clusters as distinct contexts. We conduct multiple experiments by varying the number of contexts $K$, using values of 2, 3, 4, 6, and 8. For a fair comparison, we maintain consistency in the number of contexts across different methods, ensuring that the same $K$ value corresponds to the number of mixture components in MixNorm and the number of modes in ModeNorm. The models are trained for 200 epochs with a batch size of 64, utilizing Nesterov's accelerated gradient~\cite{bengio2013advances}. The learning rate is initially set to 0.1 and is reduced by a factor of 10 at 50\% and 75\% of the total training epochs. Additionally, weight decay is fixed at $10^{-4}$ and momentum at 0.9.\newline
\begin{table}[!htbp]
    \centering
    \begin{tabular}{lllll}
       \hline
        \textbf{method}  & \textbf{CIFAR-10} & \textbf{Oxford-IIIT Pet} & \textbf{CIFAR-100} & \textbf{Tiny ImageNet} \\
        \hline
        BN & 92.07 & 75.63 & 71.35 & 52.20  \\
        LN &  84.65 & 66.12 & 58.34 & 47.20  \\
        \hline
        MixNorm-2 & 93.10 & 74.34 & 73.23 & 53.20  \\
        MixNorm-4 & 93.60 & 75.67 & 73.40 & 53.24  \\
        MixNorm-6 & 93.60 & 75.65 & 73.47 & 53.18  \\
        MixNorm-8 & 92.62 & 75.80 & 73.47 & 53.67  \\
        \hline
        ModeNorm-2 & 93.32 & 75.87 & 72.90 & 53.16  \\
        ModeNorm-4 & 93.65 & 75.84 & 73.43 & 54.12  \\
        ModeNorm-6 & 93.68 & 75.97 & 73.45 & 54.18  \\
        ModeNorm-8 & 93.68 & 76.02 & 73.27 & 54.18 \\
        \hline
        CN-2 & 93.87 & 75.98 & 73.88 & 54.15  \\
        CN-4 & 93.98 & 76.12 & 74.10 & 54.21  \\
        CN-6 & 93.98 & 76.22 & 74.10 & 54.30  \\
        CN-8 & 94.01 & 76.37 & 74.12 & 54.30  \\
        \hline
        CN-X-2 & 94.06 & 75.34 & 73.99 & 54.23  \\
        CN-X-4 & 94.05 & 76.23 & 74.34 & 55.12  \\
        CN-X-6 & 94.13 & 76.35 & 74.23 & 55.09  \\
        CN-X-8 & 94.54 & 76.35 & 74.78 & 55.26  \\
        \hline
        ACN-2 & 92.65 & 75.76 & 73.77 & 53.98  \\
        ACN-4 & 93.67 & 75.87 & 73.88 & 54.01  \\
        ACN-6 & 93.89 & 75.90 & 74.01 & 54.23  \\
        ACN-8 & 94.13 & 75.90 & 74.01 & 54.36  \\
        \hline
    \end{tabular}
    \caption{Performance (accuracy \%) of DenseNet-40 on CIFAR-10, Oxford-IIIT Pet, CIFAR-100, and Tiny ImageNet. Contexts for CN and CN-X are built using k-means clusters. "2, 3, 4, 8" represent mixture components, modes, and contexts for MixNorm, ModeNorm, and the proposed CN, CN-X, and ACN methods.}
    \label{densenet-40-1}
\end{table}
\begin{table}[!htbp]
    \centering
    \begin{tabular}{lllll}
       \hline
        \textbf{method}  & \textbf{CIFAR-10} & \textbf{Oxford-IIIT Pet} & \textbf{CIFAR-100} & \textbf{Tiny ImageNet} \\
        \hline
        BN & 94.10 & 76.28 & 73.32 & 55.12  \\
        LN &  85.20 & 66.34 & 60.10 & 47.53  \\
        \hline
        MixNorm-2 & 94.54 & 76.67 & 74.12 & 55.67  \\
        MixNorm-4 & 94.56 & 76.73 & 74.32 & 55.56  \\
        MixNorm-6 & 94.56 & 76.75 & 74.67 & 55.70  \\
        MixNorm-8 & 95.01 & 76.87 & 74.72 & 55.74  \\
        \hline
        ModeNorm-2 & 94.65 & 76.87 & 74.21 & 54.76  \\
        ModeNorm-4 & 94.67 & 76.84 & 74.34 & 55.01  \\
        ModeNorm-6 & 94.74 & 76.89 & 74.52 & 55.12  \\
        ModeNorm-8 & 94.74 & 76.89 & 74.57 & 55.12 \\
        \hline
        CN-2 & 95.10 & 76.12 & 74.67 & 55.26  \\
        CN-4 & 95.76 & 76.92 & 74.72 & 55.17  \\
        CN-6 & 95.76 & 76.92 & 74.77 & 55.78  \\
        CN-8 & 95.67 & 76.93 & 74.77 & 55.98  \\
        \hline
        CN-X-2 & 95.56 & 76.67 & 75.01 & 55.23  \\
        CN-X-4 & 95.76 & 76.87 & 75.10 & 55.76  \\
        CN-X-6 & 95.87 & 76.87 & 75.10 & 55.78  \\
        CN-X-8 & 96.12 & 77.01 & 75.21 & 55.97  \\
        \hline
        ACN-2 & 94.76 & 76.67 & 74.78 & 55.22  \\
        ACN-4 & 94.76 & 76.87 & 74.88 & 55.43  \\
        ACN-6 & 94.87 & 76.89 & 75.10 & 55.88  \\
        ACN-8 & 95.10 & 76.89 & 75.21 & 55.89  \\
        \hline
    \end{tabular}
    \caption{Performance (accuracy \%) of DenseNet-100 on CIFAR-10, Oxford-IIIT Pet, CIFAR-100, and Tiny ImageNet. Contexts for CN and CN-X are built using k-means clusters. "2, 4, 6, 8" represent mixture components, modes, and contexts for MixNorm, ModeNorm, and the proposed CN, CN-X, and ACN methods.}
    \label{densenet-100-1}
\end{table}
Table~\ref{densenet-40-1} presents the performance comparison of CN, CN-X, and ACN on a shallow neural network (DenseNet-40), while Table~\ref{densenet-100-1} highlights their effectiveness on a deeper network (DenseNet-100). Across all datasets, which vary in complexity based on the number of classes, our proposed method consistently achieves higher average accuracy. This improvement is particularly noticeable with CN-X. Additionally, when varying the number of contexts (2, 4, 6, and 8), the performance difference remains minimal, suggesting that a large number of clusters is not necessary to achieve optimal performance.
\begin{figure*}[!htbp]
    \centering
    \begin{subfigure}[b]{0.23\textwidth}
        \centering
        \includegraphics[width=\textwidth]{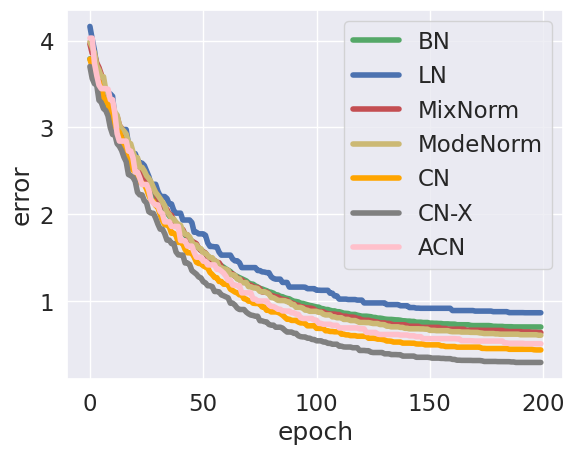}
        \caption{CIFAR-10}
        \label{fig:bn}
    \end{subfigure}
    \hfill
    \begin{subfigure}[b]{0.23\textwidth}
        \centering
        \includegraphics[width=\textwidth]{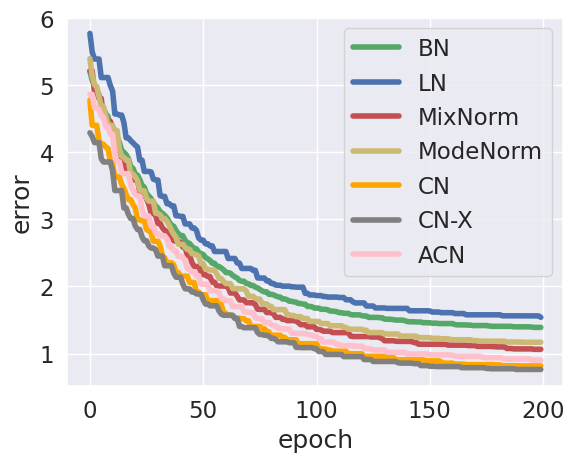}
        \caption{Oxford-IIIT Pet}
        \label{fig:bn}
    \end{subfigure}
    \hfill
    \begin{subfigure}[b]{0.23\textwidth}
        \centering
        \includegraphics[width=\textwidth]{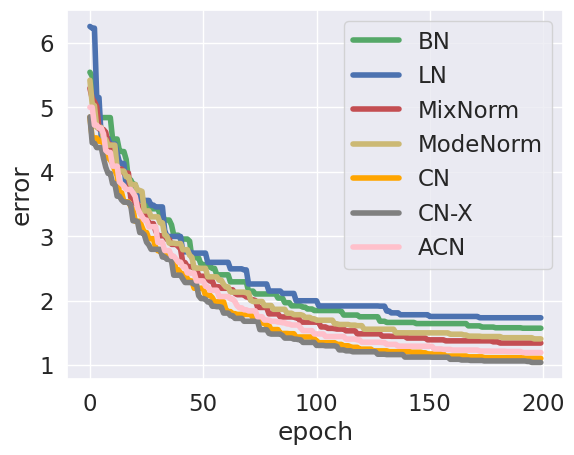}
        \caption{CIFAR-100}
        \label{fig:CN-Patches}
    \end{subfigure}
    \hfill
    \begin{subfigure}[b]{0.23\textwidth}
        \centering
        \includegraphics[width=\textwidth]{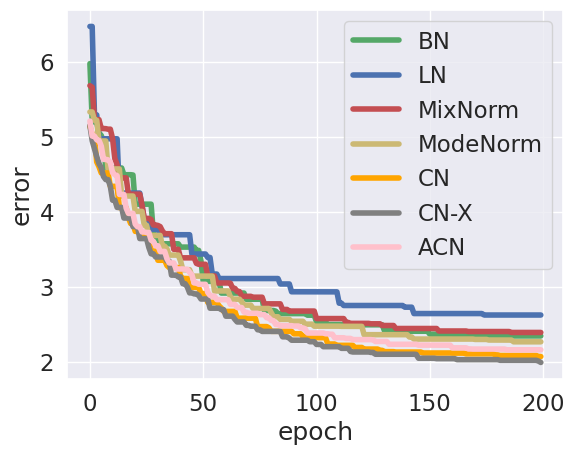}
        \caption{Tiny ImageNet}
        \label{fig:CN-Channels}
    \end{subfigure}
    \caption{DenseNet-40}
    \label{fig:cifar100_loss}


    \begin{subfigure}[b]{0.23\textwidth}
        \centering
        \includegraphics[width=\textwidth]{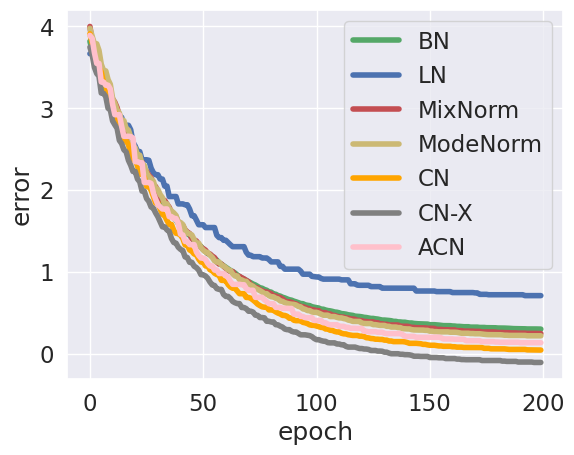}
        \caption{CIFAR-10}
        \label{fig:figure1}
    \end{subfigure}
    \hfill
    \begin{subfigure}[b]{0.23\textwidth}
        \centering
        \includegraphics[width=\textwidth]{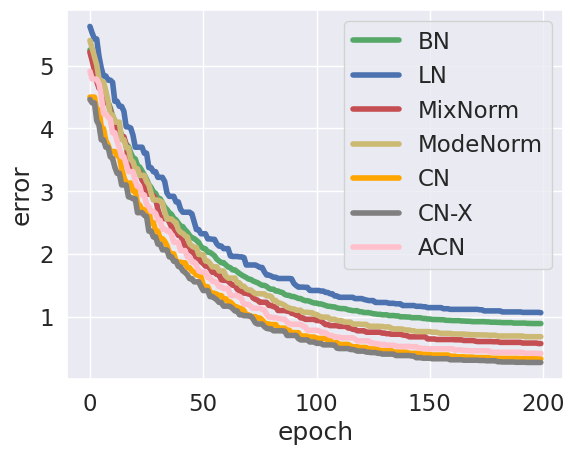}
        \caption{Oxford-IIIT Pet}
        \label{fig:figure1}
    \end{subfigure}
    \hfill
    \begin{subfigure}[b]{0.23\textwidth}
        \centering
        \includegraphics[width=\textwidth]{images/densenet100_cifar10.png}
        \caption{CIFAR-100}
        \label{fig:figure2}
    \end{subfigure}
    \hfill
    \begin{subfigure}[b]{0.23\textwidth}
        \centering
        \includegraphics[width=\textwidth]{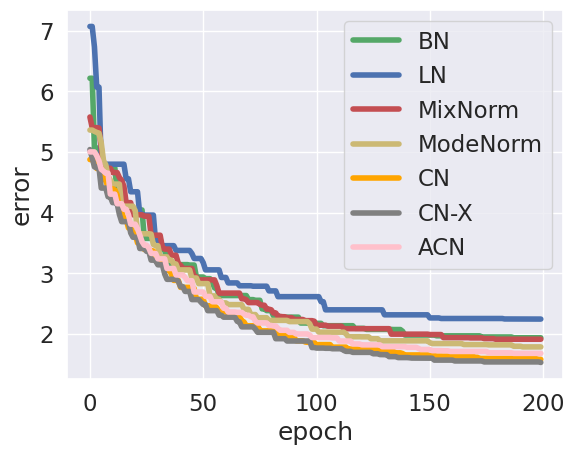}
        \caption{Tiny ImageNet}
        \label{fig:figure3}
    \end{subfigure}
    \caption{DenseNet-100}
    \label{fig:second_set_of_figures}
    \caption{Training Error Trends for DenseNet-40 and DenseNet-100 with Various Normalization Layers. The MixNorm, ModeNorm, CN, CN-X, and ACN methods are implemented using $K=8$.}
    \label{fig:all_figures}
\end{figure*}
Figure~\ref{fig:all_figures} demonstrates that CN, CN-X, and ACN achieve superior convergence compared to traditional methods such as BN, LN, MixNorm, and ModeNorm. The observed acceleration in convergence, illustrated in Figure~\ref{fig:all_figures}, alongside the improved performance metrics presented in Tables~\ref{densenet-40-1} and~\ref{densenet-100-1}, indicates that our proposed method can effectively serve as a layer to enhance model performance and accelerate convergence, even when prior knowledge of the datasets is limited. In such cases, the k-means algorithm can be employed to generate clusters, which can then be used as contexts for CN and CN-X.\newline
Conversely, when we have a thorough understanding of the dataset and the contexts are well-defined, there is no need to apply k-means clustering; instead, we can directly utilize predefined contexts. This approach will be elaborated upon in the following section.

\subsubsection{Leveraging Predefined Contexts}
\label{predefined_context}
Some datasets, such as Oxford-IIIT Pet and CIFAR-100, not only have a hierarchical structure of classes but also include superclasses that group similar classes together. For instance, in the Oxford-IIIT Pet dataset, various breeds of dogs and cats can be categorized into two superclasses: "dog" and "cat". Similarly, CIFAR-100 contains 20 distinct superclasses. Rather than applying the k-means algorithm to create clusters for use as contexts, we can leverage these existing superclasses as contextual representations.\newline
In this experiment, we employ the same models as in the previous section, specifically DenseNet-40 and DenseNet-100, to evaluate the evolution of accuracy on the CIFAR-100 and Oxford-IIIT Pet datasets. We utilize the superclasses as contexts and implement normalization layers CN, CN-X, and ACN. The goal is to assess whether a deeper understanding of our dataset, achieved by constructing contexts, yields improved performance compared to relying on predefined contexts (superclasses) present in the datasets.
\begin{table*}[!htbp]
    \centering
    \begin{tabular}{lllllll}
        \hline
        \multicolumn{7}{c}{\textbf{Oxford-IIIT Pet}}\\
       \hline
        model   & 25 epochs & 50 epochs & 75 epochs & 100 epochs & 150 epochs & 200 epochs \\
        \hline
        CN & 75.43 & 76.86 & 76.88 & 77.34 & 78.43 & 79.26 \\
        CN-X & 76.12 & 76.77  & 77.98 & 78.66 & 80.02 & 80.98 \\
        ACN & 72.34 & 72.56 & 73.10 & 74.22 & 74.90 & 76.13 \\
        \hline
    \end{tabular}
    
    \begin{tabular}{lllllll}
        \hline
        \multicolumn{7}{c}{\textbf{CIFAR-100}}\\
       \hline
        model   & 25 epochs & 50 epochs & 75 epochs & 100 epochs & 150 epochs & 200 epochs \\
        \hline
        CN & 73.88 & 74.21 & 74.89  & 75.10 & 76.53 & 77.67 \\
        CN-X & 74.21 & 75.10 & 75.67 & 77.45 & 78.54 & 79.78 \\
        ACN & 72.34 & 72.67 & 74.32 & 74.32 & 74.56 & 74.60 \\
        \hline
    \end{tabular}

    \caption{Evolution of Accuracy with DenseNet-40 Utilizing Superclasses as Contexts on the Oxford-IIIT Pet and CIFAR-100 Datasets.}
   \label{table:densenet-40-superclasses}
\end{table*}
\begin{table*}[!htbp]
    \centering
    \begin{tabular}{lllllll}
        \hline
        \multicolumn{7}{c}{\textbf{Oxford-IIIT Pet}}\\
       \hline
        model   & 25 epochs & 50 epochs & 75 epochs & 100 epochs & 150 epochs & 200 epochs \\
        \hline
        CN & 75.43 & 75.67 & 76.98 & 77.89 & 79.34 & 80.23 \\
        CN-X & 76.54 & 77.87 & 79.78 & 81.23 & 81.23 & 82.02  \\
        ACN & 73.02 & 74.32 & 75.43 & 77.02 & 77.32 & 77.85 \\
        \hline
    \end{tabular}
    
    \begin{tabular}{lllllll}
        \hline
        \multicolumn{7}{c}{\textbf{CIFAR-100}}\\
       \hline
        model   & 25 epochs & 50 epochs & 75 epochs & 100 epochs & 150 epochs & 200 epochs \\
        \hline
        CN & 74.21 & 74.56 & 76.78 & 78.22 & 78.22 & 79.34 \\
        CN-X & 73.56 & 75.43 & 75.78 & 79.34 & 79.89 & 81.02 \\
        ACN & 73.21 & 73.76 & 75.11 & 76.21 & 76.21 & 76.32 \\
        \hline
    \end{tabular}

    \caption{ Evolution of Accuracy with DenseNet-100 Utilizing Superclasses as Contexts on the Oxford-IIIT Pet and CIFAR-100 Datasets.}
   \label{table:densenet-100-superclasses}
\end{table*}
Tables~\ref{table:densenet-40-superclasses} and~\ref{table:densenet-100-superclasses} illustrate the significant impact that well-defined contexts have on the performance of CN and CN-X. Notably, when utilizing superclasses as contexts, we achieve comparable performance in approximately 25 epochs, in contrast to the 200 epochs required when using k-means clusters, as detailed in Tables~\ref{densenet-40-1} and~\ref{densenet-100-1}. Furthermore, employing $K=2$ for the Oxford-IIIT Pet dataset and $K=20$ for CIFAR-100 does not markedly affect ACN performance. This suggests that since contexts are constructed within ACN, merely increasing the number of contexts does not guarantee enhanced model performance.\newline
This experiment highlights the potential advantages of applying CN and CN-X for normalization when we possess a strong understanding of the datasets, allowing us to leverage this knowledge as prior information to construct effective contexts that yield improved performance in both shallow and deep neural networks.\newline
To further evaluate the versatility of CN, CN-X, and ACN, we implement the Vision Transformer (ViT) model~\cite{dosovitskiy2020image} and compare its performance against BN, LN, MixNorm, and ModeNorm on the CIFAR-100 dataset. For CN and CN-X, we utilize superclasses as contexts with $K=20$. In the case of ACN, ModeNorm, and MixNorm, we also set $K=20$ to ensure a fair comparison across all methods.
\begin{table}[!htbp]
    \centering
    \begin{tabular}{llllll}
        \hline
        model & accuracy & precision & recall & f1-score \\
        \hline
        BN & 55.63 &  8.96 & 90.09  & 54.24 \\
        LN & 54.05  & 11.82 & 85.05 & 53.82 \\
        MixNorm & 53.2 & 11.20 & 87.10 & 54.23 \\
        ModeNorm & 54.10 & 12.12 & 87.23 & 54.98 \\
        \hline
        CN & 70.76 & 27.59 & 98.60 & 70.70 \\
        CN-X & 71.28 & 28.30 & 98.87 & 70.98 \\
        ACN & 60.34 & 20.21 & 93.23 & 60.10 \\
        \hline
    \end{tabular}
    \caption{Performance Rates (\%) on the Test Set Using the ViT Architecture with Various Normalization Methods—BN, LN, MixNorm, ModeNorm, CN, CN-X, and ACN—on the CIFAR-100 Dataset, Employing Superclasses as Contexts for CN and CN-X.}
    \label{table:cifar_superclass}
\end{table}
\begin{figure*}[!htbp]
    \centering
    \begin{subfigure}{0.45\textwidth}
        \centering
        \includegraphics[width=\textwidth]{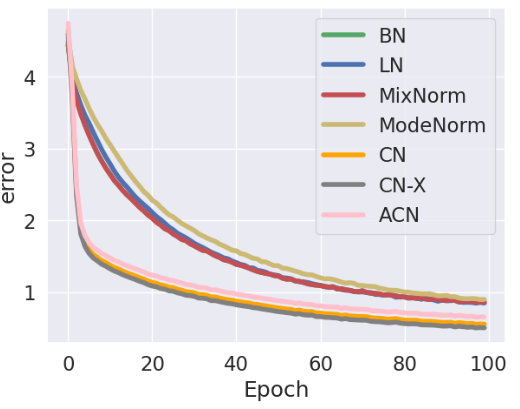}
        \caption{Training Error}
        \label{fig:figure1}
    \end{subfigure}
    \hspace{15pt}
    \begin{subfigure}{0.45\textwidth}
        \centering
        \includegraphics[width=\textwidth]{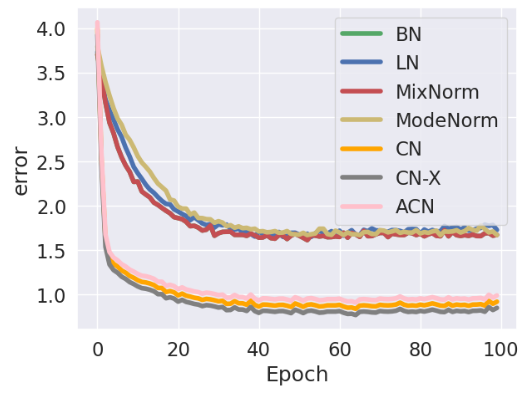}
        \caption{Validation Error}
        \label{fig:figure2}
    \end{subfigure}
    \caption{Contrasting Training and Validation Error Curves in CIFAR-100 dataset when using ViT architecture.}
    \label{fig:overall}
\end{figure*}
Table~\ref{table:cifar_superclass} demonstrates the versatility of the proposed normalization methods CN, CN-X, and ACN. When applied to the ViT architecture, these methods maintain a performance advantage over BN, LN, MixNorm, and ModeNorm. Similarly to the results obtained with DenseNet, the proposed normalization layers facilitate improved convergence during training and validation, as illustrated in Figure~\ref{fig:overall}.\newline
In this section, we demonstrate that our proposed normalization methods significantly enhance performance and accelerate convergence in both shallow and deep neural networks. When predefined contexts are not available, we illustrate the feasibility of using k-means clusters as an alternative. Conversely, when contexts are well-defined—such as through superclasses for CN and CN-X—we achieve improved performance. We provide evidence of this through applications with CNN architectures, specifically DenseNet-40 and DenseNet-100, as well as with the Transformer architecture using ViT~\cite{vaswani2017attention}.\newline
To further explore these findings, we propose an additional approach in the following section to effectively construct contexts for CN and CN-X, demonstrating the versatility of these methods and their applicability across various domains.

\subsection{Domain Adaptation}
\label{domain_adaptation}
In this experiment, we introduce an alternative approach to constructing contexts for CN and CN-X in domain adaptation. Domain adaptation~\cite{farahani2021brief} is a technique in machine learning, particularly in deep learning, that enables a model trained on data from one domain (source domain) to perform well on data from a different but related domain (target domain). This is useful when labeled data is abundant in the source domain but limited or unavailable in the target domain, which may have different characteristics, like variations in lighting, style, or noise. By aligning feature distributions or representations between domains, domain adaptation allows the model to generalize better across domains, improving performance on tasks where collecting labeled data is challenging.\newline
For CN and CN-X, we will consider two distinct contexts $K=2$: the source domain and the target domain. Using domains as contexts is motivated by the aim to incorporate domain-specific information into the activation representations. To exemplify this, we employ AdaMatch~\cite{berthelot2021adamatch}, a domain adaptation algorithm designed to align feature distributions between source and target domains by leveraging labeled source data and a few labeled target samples. AdaMatch uses a dynamically adjusted confidence threshold for pseudo-labeling in the target domain, improving generalization across domains by aligning class distributions while minimizing domain shift. It combines the tasks of unsupervised domain adaptation (UDA), semi-supervised learning (SSL), and semi-supervised domain adaptation (SSDA). In UDA, we have access to a labeled dataset from the source domain and an unlabeled dataset from the target domain, with the goal of training a model that generalizes effectively to the target data. In this case, we use MNIST as the source dataset and SVHN as the target dataset. These datasets include a range of variations, such as texture, viewpoint, and appearance, and their respective domains, or distributions, are notably distinct.\newline
The baseline model uses BN layers and is trained from scratch using Wide Residual Networks~\cite{zagoruyko2016wide}. For comparison, we create additional models by individually replacing the BN layers with LN, MixNorm, ModeNorm, CN, CN-X, and ACN. For MixNorm, ModeNorm, and ACN, we set $K=2$ to maintain consistency with CN and CN-X. Model training employs the Adam optimizer~\cite{kingma2014adam} with a cosine decay schedule, gradually reducing the initial learning rate of $0.03$. All models are trained for 100 epochs.
\begin{table}[!htbp]
    \centering
    \begin{tabular}{llllll}
        \hline
        \multicolumn{5}{c}{\textbf{MNIST (source domain)}}\\
        \hline
        model  & accuracy & precision & recall & f1-score  \\
        \hline
        BN & 97.36 & 87.33 & 79.39 & 78.09 \\
        LN & 96.23 & 88.26 & 76.20 & 81.70 \\
        MixNorm & 98.90 & 98.45 & 98.89 & 98.93 \\
        ModeNorm & 98.93 & 98.3 & 98.36 & 98.90 \\
        \hline
        CN & 99.17 & 99.17 & 99.17 & 99.17 \\
        CN-X & 99.26 & 99.20 & 99.32 & 99.26 \\
        ACN & 98.9 & 98.5 & 98.90 & 98.95 \\
        \hline
    \end{tabular}
    \begin{tabular}{llllll}
        \hline
        \multicolumn{5}{c}{\textbf{SVHN (target domain)}}\\
        \hline
        model  & accuracy & precision & recall & f1-score  \\
        \hline
        BN & 25.08 & 31.64 & 20.46 & 24.73 \\
        LN & 24.10 & 28.67 & 22.67 &  23.67 \\
        MixNorm & 32.14 & 50.12 & 37.14 & 39.26 \\
        ModeNorm & 32.78 & 49.87 & 38.13 & 40.20 \\
        \hline
        CN & 47.63 & 60.90 & 47.63 & 49.50 \\
        CN-X & 54.70 & 59.74 & 54.70 & 54.55 \\
        ACN & 33.4 & 43.83 & 40.28 & 42.87 \\
        \hline
    \end{tabular}
    \caption{Test set accuracy (\%) of AdaMatch for domain adaptation on MNIST and SVHN datasets using various normalization techniques.}
    \label{table:domain_adaptation}
\end{table}
The results in Table~\ref{table:domain_adaptation} demonstrate that CN, CN-X, and ACN outperform traditional normalization techniques (BN, LN, MixNorm, and ModeNorm) in domain adaptation between MNIST and SVHN. For the MNIST source domain, all methods achieve high performance, with CN-X achieving the best accuracy and F1-score of 99.26\%. In contrast, performance differences are more pronounced on the SVHN target domain, where CN-X leads with a significant improvement in accuracy (54.70\%), followed closely by CN at 47.63\%. These results suggest that CN and CN-X are better suited to handle domain shifts, particularly when there is a substantial difference in data distribution, as seen between MNIST and SVHN. While ACN does not reach the peak accuracy levels of CN-X on SVHN, it still shows a marked improvement over baseline methods like BN and LN, achieving 33.4\% accuracy in the target domain. This indicates that ACN contributes to enhanced domain adaptation by capturing some domain-specific features, making it a viable normalization technique for adaptation tasks, though its performance suggests it is less robust to drastic domain shifts compared to CN and CN-X.\newline
These results from CN and CN-X reinforce findings from previous experiments, where contexts are clearly defined. Leveraging well-defined prior knowledge can be highly beneficial, as it allows relevant patterns to be embedded within activation representations. This enhances the overall representation quality and provides normalization benefits that contribute to the stability of the training process. By capturing domain-specific information effectively, CN and CN-X not only improve adaptation to new domains but also support smoother learning by reducing the impact of domain shifts on model performance. This approach highlights the potential of context-driven normalization techniques to boost model robustness in challenging cross-domain tasks, as seen with AdaMatch on the MNIST to SVHN adaptation.\newline
In the next section, we will examine a scenario where the application of ACN is particularly relevant and compare its performance to single-mode normalization (BN) and multi-mode normalization (MixNorm).

\subsection{Image Generation}
\label{image_generation}
Image generation involves creating new, synthetic images by training models to understand and replicate the features and patterns of real images. This process uses a model to learn from a large dataset of images, capturing details like textures, colors, shapes, and spatial relationships. Generated images can range from realistic representations to imaginative interpretations, depending on the training data and model design. An example of method that can generate such images is Generative Adversarial Networks (GANs)~\cite{radford2015unsupervised, ledig2017photo, isola2017image}. The GAN architecture consists of two neural networks: a generator and a discriminator, which work in tandem through a process called adversarial training. The generator creates synthetic images starting from random noise, while the discriminator evaluates these images, distinguishing between real images (from the training dataset) and those generated by the model. The generator's goal is to create images that can "fool" the discriminator, while the discriminator aims to accurately detect real versus generated images. This adversarial process continues until the generator produces images that are nearly indistinguishable from real ones.
GANs have a wide range of applications, including image synthesis, style transfer, super-resolution imaging, and data augmentation. They are also used in fields like healthcare for generating medical images, in entertainment for creating realistic character images, and in autonomous driving for simulating varied road conditions. A common challenge encountered when using GANs is the issue of "mode collapse". This phenomenon occurs when the generator produces only a restricted subset of possible data, leading to a loss of diversity in the generated results. In other words, the generator may focus on producing a specific type of data, neglecting the generation of other potential variations. This problem can compromise the quality and variety of the generated data, requiring specific techniques and strategies to address and enhance the overall performance of the GAN model. In MixNorm~\cite{kalayeh2019training}, the authors demonstrate that normalizing across multiple modes (mixture components), rather than a single mode as in BN, can help mitigate the issue of "mode collapse". Here, we propose to apply ACN and compare its performance to BN and MixNorm. Notably, CN and ACN are not suited for this scenario, as generated images are produced from random noise vectors, making it difficult to define prior knowledge about vector membership for normalization.\newline
Our baseline model is a Deep Convolutional Generative Adversarial Network (DCGAN)~\cite{radford2015unsupervised}, specifically designed for image generation. The generator consists of a linear layer followed by four deconvolutional layers, with the first three layers utilizing Batch Normalization (BN) and a LeakyReLU~\cite{maas2013rectifier} activation function. The linear layer maps latent space to a higher-dimensional representation, while the deconvolutional layers progressively upsample the input into realistic images. BN stabilizes and accelerates training, and LeakyReLU introduces non-linearity for better learning of complex mappings. We create two additional models by replacing the BN layers with MixNorm and ACN, using $K=3$ for MixNorm as specified in the paper~\cite{kalayeh2019training} and matching $K=3$ for ACN to ensure a fair comparison. All models are trained on CIFAR-100 for 200 epochs using the Adam optimizer~\cite{kingma2014adam} with $\alpha=0.0002$, $\beta_1=0$, and $\beta_2=0.9$ for both the generator and discriminator. We evaluate GAN quality using the Fréchet Inception Distance (FID)~\cite{heusel2017gans}, calculated every 10 epochs for efficiency.
\begin{figure*}[!htbp]
    \centering
    \includegraphics[width=0.5\textwidth]{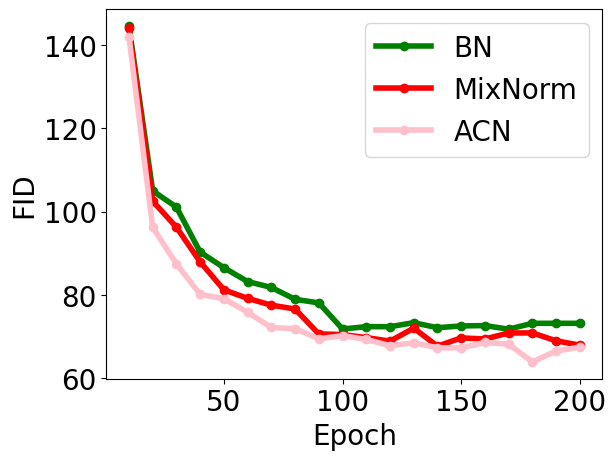}
    \caption{ACN integrated as a normalization layer in a DCGAN. Our results show that incorporating ACN into the DCGAN generator leads to improved (lower) Fréchet Inception Distance (FID) scores.}
    \label{fig:fid}
\end{figure*}
Figure~\ref{fig:fid} illustrates that the DCGAN incorporating ACN exhibits not only a quicker convergence compared to its batch-normalized (BN) and mixture-normalized (MixNorm) counterparts but also achieves superior (lower) FID scores. Reducing the FID is crucial as it indicates that the generated images are more similar to real images, enhancing the overall quality and diversity of the outputs. A lower FID score suggests that the model is effectively capturing a broader range of features in the training data, which helps mitigate mode collapse—a phenomenon where the generator produces a limited variety of outputs. By improving the distribution of generated images and reducing the gap between real and synthetic data distributions, ACN promotes a more stable training process and encourages the model to explore different modes within the data, leading to richer and more varied image generation.
\begin{figure*}[!htbp]
    \centering
    \begin{subfigure}[b]{0.30\textwidth}
        \centering
        \includegraphics[width=\textwidth]{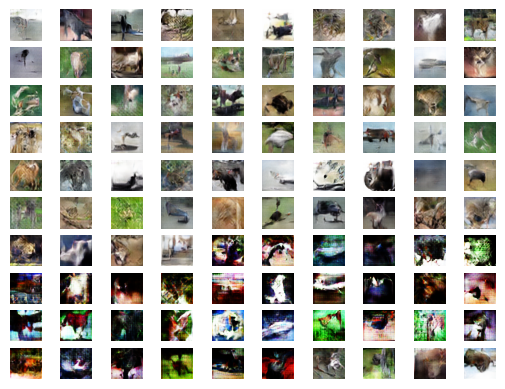}
        \caption{BN}
        \label{fig:bn_gan}
    \end{subfigure}
    \hfill
    \begin{subfigure}[b]{0.30\textwidth}
        \centering
        \includegraphics[width=\textwidth]{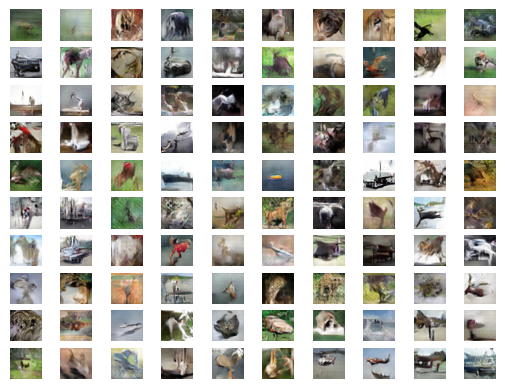}
        \caption{MixNorm}
        \label{fig:mn_gan}
    \end{subfigure}
    \hfill
    \begin{subfigure}[b]{0.30\textwidth}
        \centering
        \includegraphics[width=\textwidth]{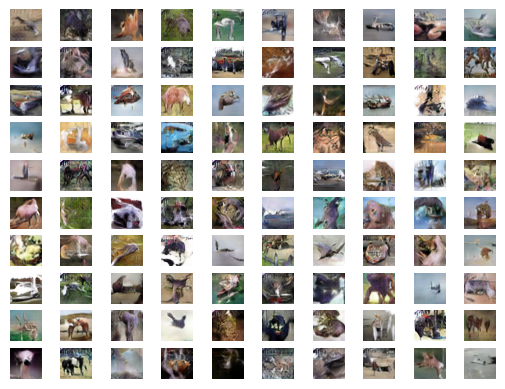}
        \caption{ACN}
        \label{fig:cn_gan}
    \end{subfigure}

    \caption{Examples of generated images at epoch $200$ are showcased for BN, MixNorm, and ACN in Figure~\ref{fig:bn_gan},\ref{fig:mn_gan}, and\ref{fig:cn_gan}, respectively.}
    \label{fig:generated}
\end{figure*}
Figure~\ref{fig:generated} showcases examples of images generated by DCGANs utilizing BN, MixNorm, and ACN. The results reveal that multi-mode normalization techniques, such as MixNorm and ACN, produce notably clearer object structures in the generated images compared to those using BN. Additionally, both MixNorm and ACN demonstrate greater diversity in their outputs, enhancing the overall richness of the generated content. This improvement in image quality and diversity underscores the effectiveness of these advanced normalization methods, paving the way for more sophisticated and nuanced image generation in future applications.\newline
\section{CONCLUSION}
\label{chapter2:discussion}
We introduce three advanced normalization techniques—Context Normalization (CN), Context Normalization Extended (CN-X), and Adaptive Context Normalization (ACN)—that aim to surpass the limitations of single-mode normalization methods like Batch Normalization and Layer Normalization. These multi-mode normalization strategies, grounded in prior knowledge, enhance activation normalization and improve the training process in neural networks.

CN and CN-X organize data into predefined groups, termed contexts, to estimate normalization parameters within each mini-batch. CN applies these parameters directly within each context, while CN-X defines them as trainable weights that are updated dynamically via backpropagation. Various methods can establish contexts, including k-means clustering, superclass assignments, and domain-based distinctions in domain adaptation. Where context construction proves challenging, ACN offers flexibility by allowing the number of contexts to be set as a hyperparameter.

Across tasks in classification, domain adaptation, and image generation, these methods consistently outperform traditional normalization techniques. CN and CN-X demonstrate strong robustness when contexts are well-defined, underscoring the impact of prior knowledge on neural network representation, faster convergence, and overall performance.

For future work, this approach may extend to multimodal representations, where structured prior knowledge could reduce parameter tuning demands and minimize reliance on large labeled datasets, driving competitive performance with fewer resources.

\bibliographystyle{ieeetr}
\bibliography{bibliography}

\clearpage

\end{document}